\documentclass[table]{ai2style/ai2}

\usepackage{microtype}
\usepackage{hyperref}
\usepackage{url}
\usepackage{booktabs, tabularx} % tabularx for ian's decon table
\usepackage{graphicx}
\usepackage{lineno}
\usepackage{enumitem}
\usepackage{listings} % for code listing
\usepackage{svg}

\definecolor{ darkblue}{rgb}{0, 0, 0.5}
\hypersetup{colorlinks=true, citecolor=darkblue, linkcolor=darkblue, urlcolor=darkblue}

\usepackage{amssymb}
\usepackage{multirow}
\usepackage{bigdelim}
\usepackage{todonotes}
\usepackage{longtable}
\usepackage{tabularray}
\usepackage{wrapfig}
\usepackage[most]{tcolorbox}
\usepackage{url}
\usepackage{xspace}
\usepackage{svg}
\usepackage[absolute]{textpos} % Add this to your preamble

\usepackage{fdsymbol}   % cute symbols

\usepackage[utf8]{inputenc} % allow utf-8 input
\usepackage[T1]{fontenc}    % use 8-bit T1 fonts
\usepackage{url}            % simple URL typesetting
\usepackage{booktabs}       % professional-quality tables
\usepackage{amsfonts}       % blackboard math symbols
\usepackage{nicefrac}       % compact symbols for 1/2, etc.
\usepackage{microtype}      % microtypography
\usepackage{amsmath}
\usepackage[most]{tcolorbox}
\usepackage{csquotes}

\usepackage{siunitx}
\usepackage{graphicx}
\usepackage{arydshln}
\usepackage{wrapfig}
\usepackage{enumitem}
\usepackage{soul} % For highlighting

\usepackage{multirow}
\usepackage{xspace}
\usepackage{adjustbox}
\usepackage{pifont}
\usepackage{caption}
\usepackage{makecell}
\usepackage{subcaption}
\usepackage{bold-extra}
\usepackage{url}
\usepackage{xcolor,colortbl}

\usepackage{pgf-pie}

\usepackage{hyperref}
\definecolor{linkcolor}{RGB}{0, 0, 128}
\hypersetup{
     colorlinks   = true,
     citecolor    = linkcolor,
     linkcolor    = linkcolor,
     urlcolor     = linkcolor,
}
\usepackage{pifont}% http://ctan.org/pkg/pifont
\usepackage{listings}
\setlist[itemize]{leftmargin=*,itemsep=0em,parsep=0.3em,topsep=0.3em}

\DeclareUnicodeCharacter{2212}{\ensuremath{-}}

\usepackage{tikz}

\usepackage{setspace}

\usepackage{nicematrix}
\newcolumntype{L}[1]{>{\raggedright\let\newline\\\arraybackslash\hspace{0pt}}m{#1}}
\newcolumntype{C}[1]{>{\centering\let\newline\\\arraybackslash\hspace{0pt}}m{#1}}
\newcolumntype{R}[1]{>{\raggedleft\let\newline\\\arraybackslash\hspace{0pt}}m{#1}}
\newcolumntype{P}[1]{>{\centering\let\newline\\\arraybackslash\columncolor{ai2lightpink}}m{#1}}
\newcommand{\aitoo}{\raisebox{-1.5pt}{\includegraphics[height=1.05em]{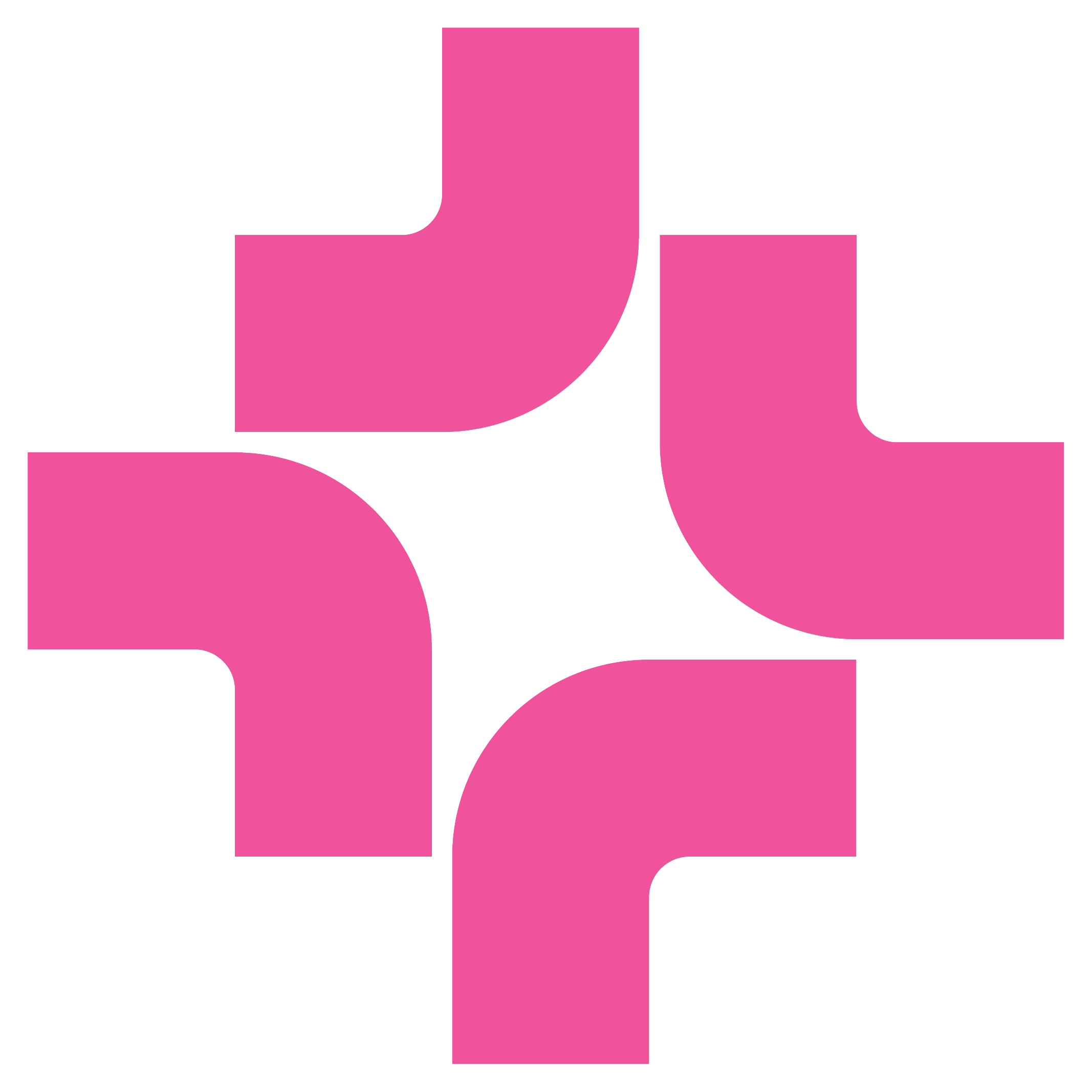}}\xspace}

\newcommand{\coreContrib}{\raisebox{.28em}{\hspace{.05em}\includegraphics[height=.45em]{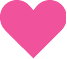}}\hspace{0.1em}}
\newcommand{\starOlmo}{\raisebox{.28em}{\hspace{.05em}\includegraphics[height=.5em]{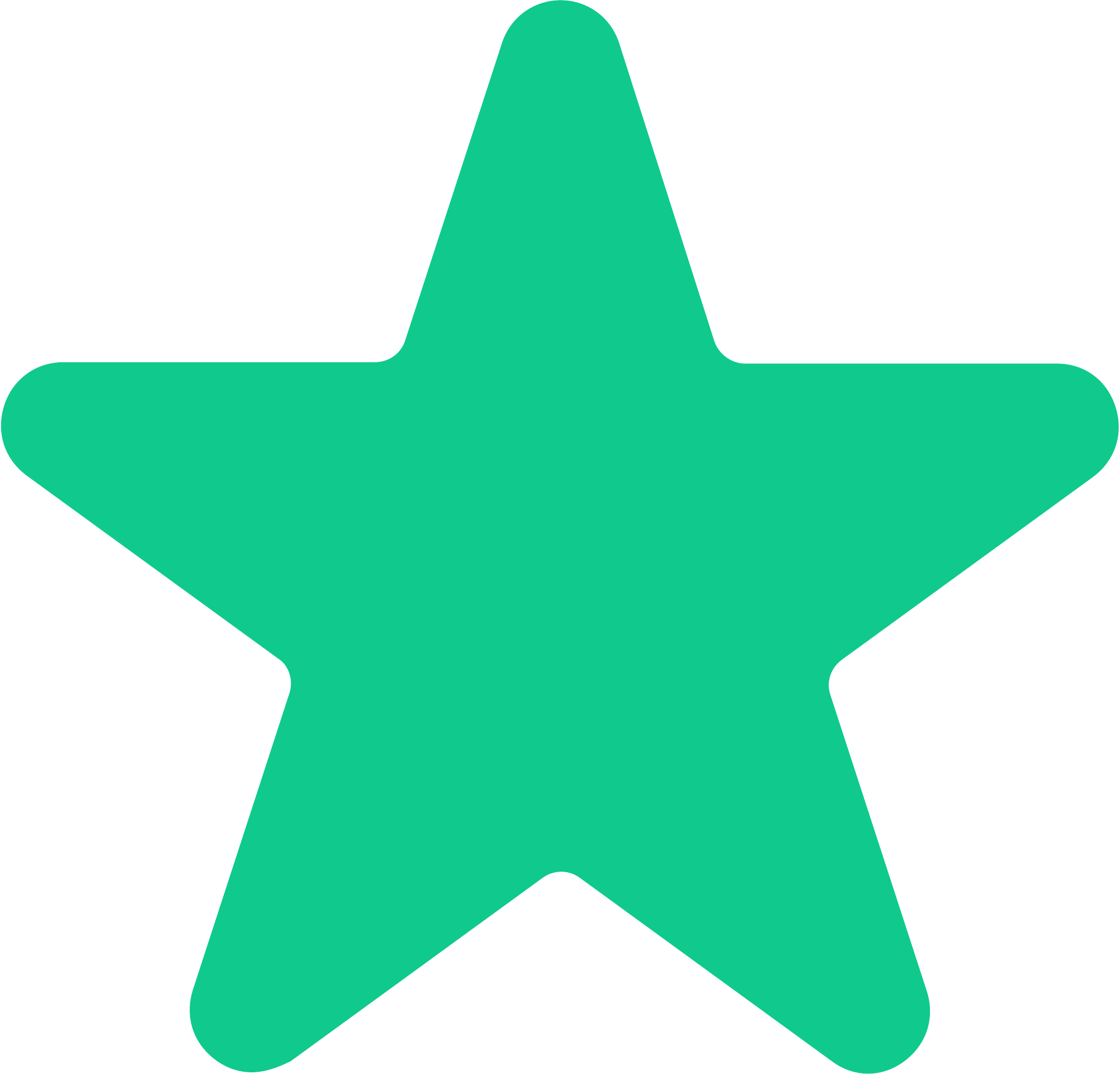}}}
\newcommand{\huggingface}{\raisebox{-1.5pt}{\includegraphics[height=1.05em]{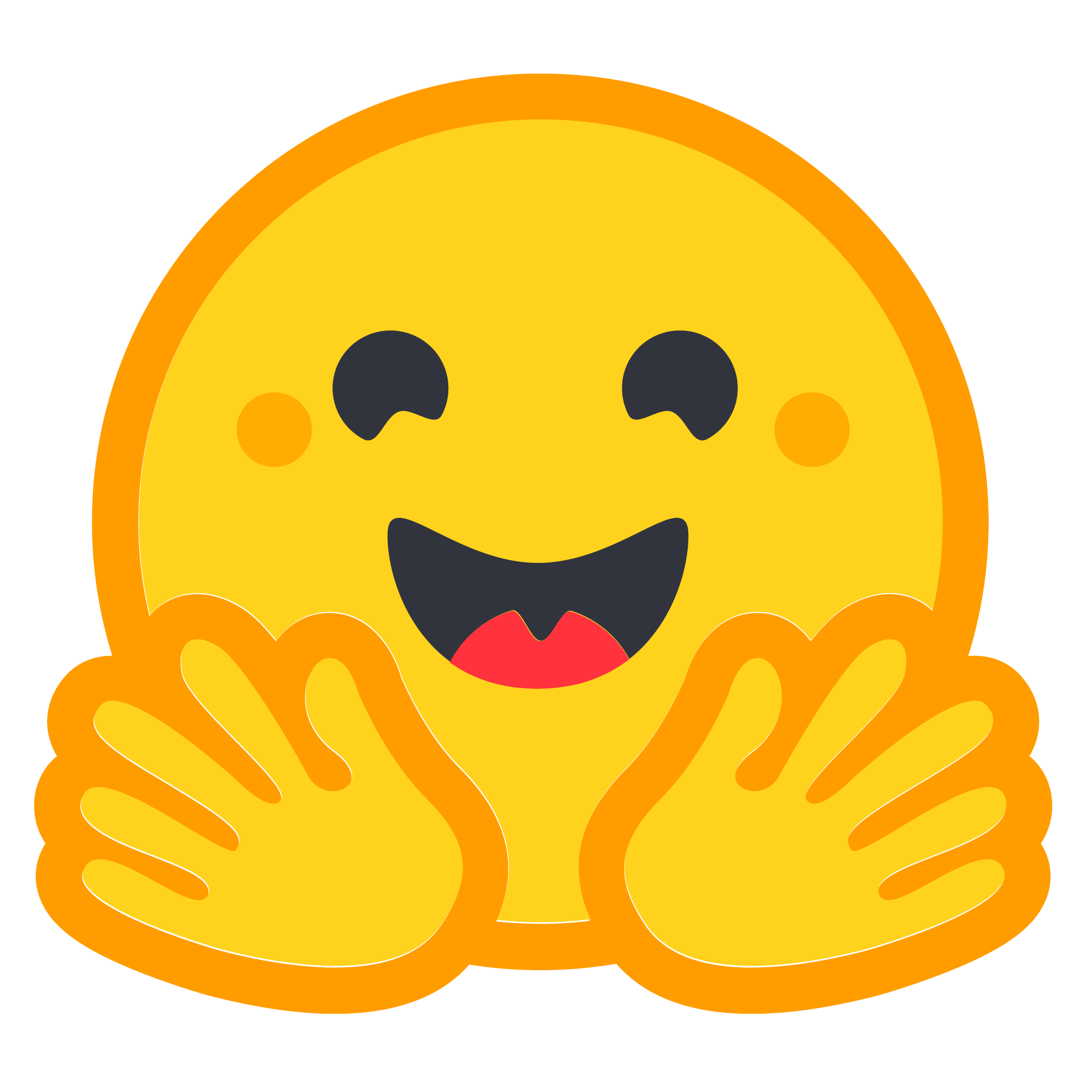}}\xspace}

\newcommand{\emailLogo}{\raisebox{-1.5pt}{\includegraphics[height=1.05em]{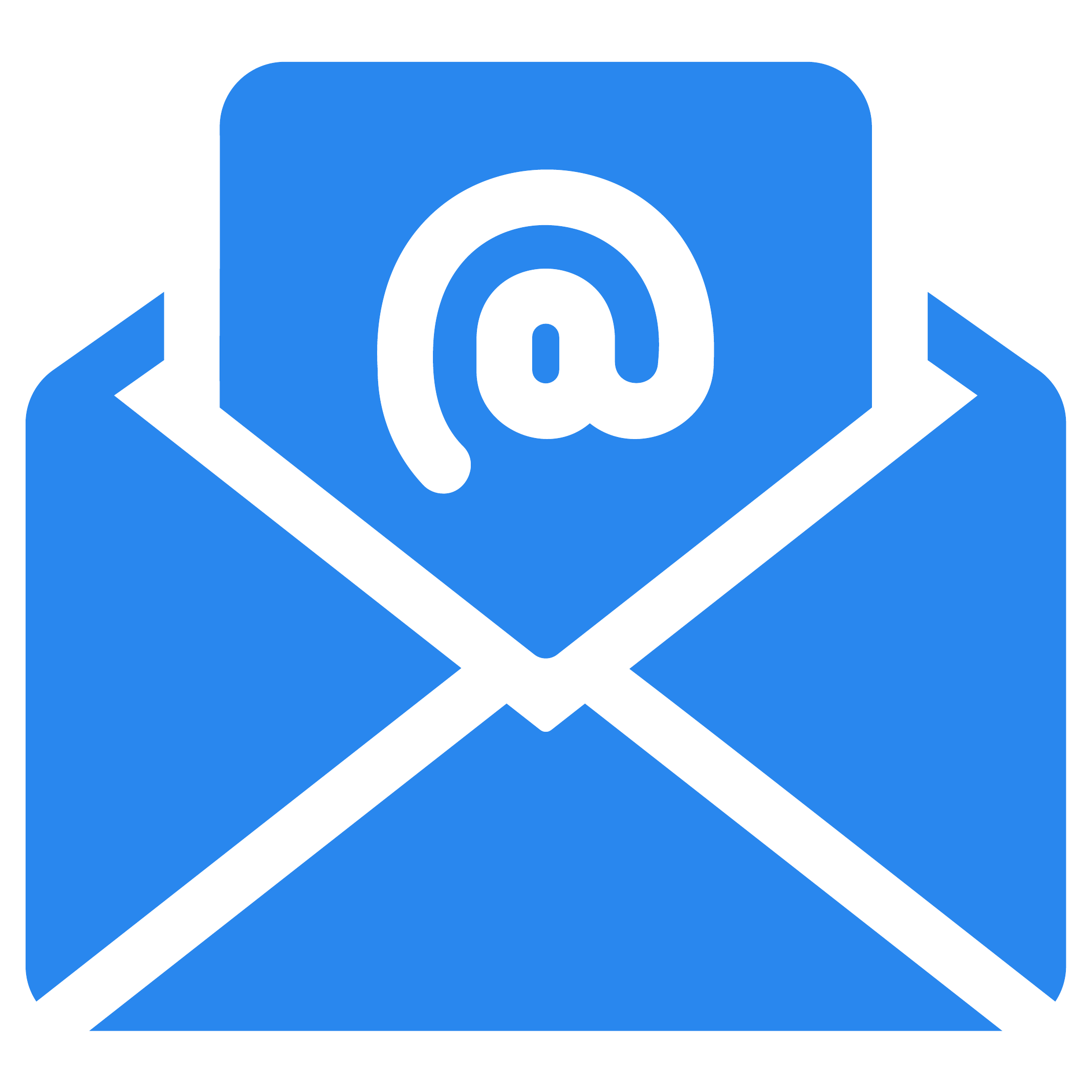}}\xspace}
\newcommand{\github}{\raisebox{-1.5pt}{\includegraphics[height=1.05em]{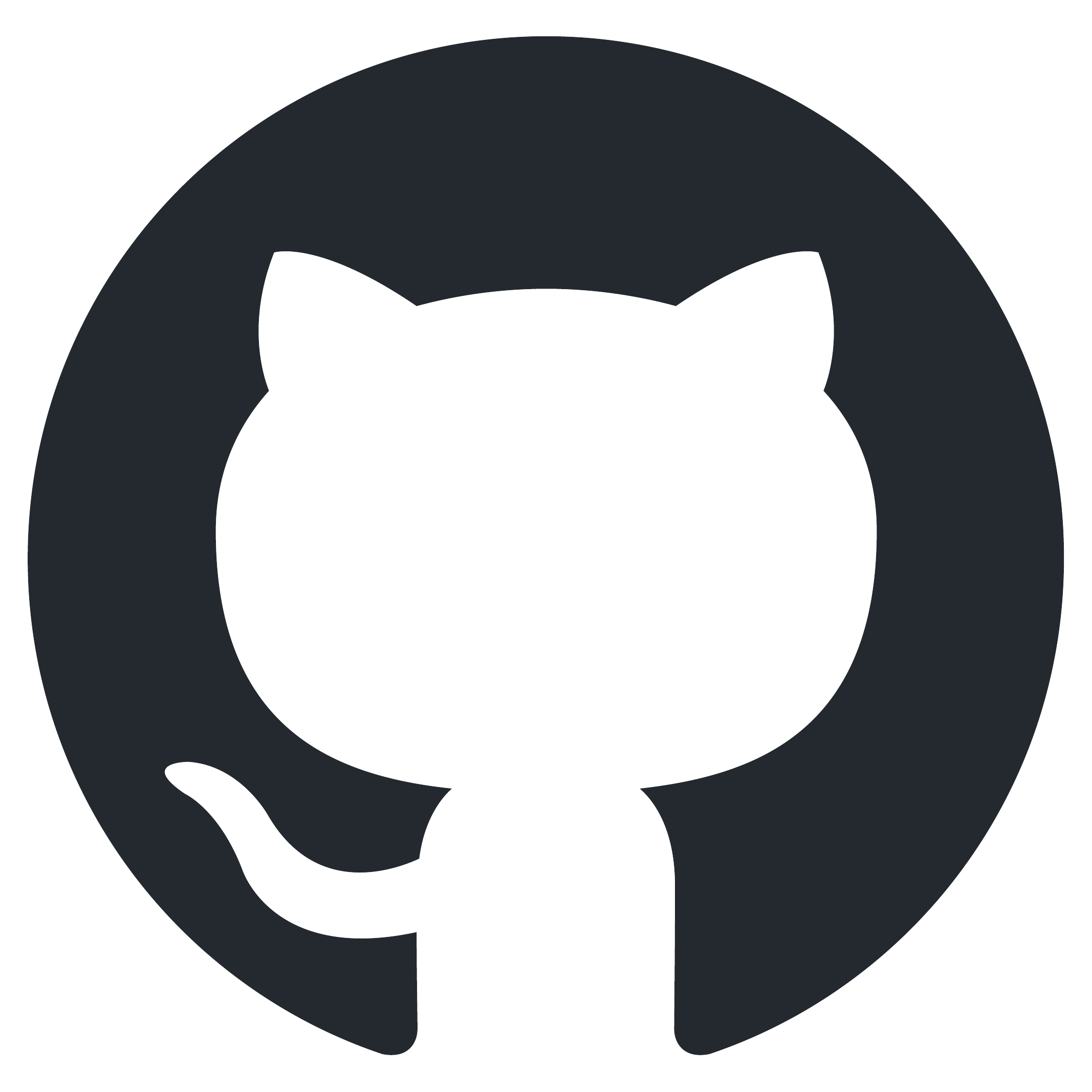}}\xspace}

\usepackage{etoolbox}
\newtoggle{anon}
\togglefalse{anon}

\def\model{OlmoEarth}

\def\modelurl{\url{https://github.com/allenai/olmoearth_pretrain}}

\def\colordefault{0a3235}
\def\colorone{f0529c}
\def\colortwo{b11be8}
\def\colorthree{6fa8dc}
\def\colorfour{0fcb8c}
\def\colorfive{12cce5}
\def\colorsix{f65834}
\def\colorgrey{B0B0B0}

\usepackage{multirow}
\usepackage{comment}
\usepackage{adjustbox}

\usepackage{tabularx}
\usepackage{graphicx}
\usepackage{tikz}
\usetikzlibrary{patterns.meta,positioning, arrows.meta,fit,calc,shapes.geometric,matrix}

\usepackage{pgfplots}
\pgfplotsset{compat=1.18}

\newcommand*{\hdr}[1]{\multicolumn{1}{l}{\rlap{\,\,\,\rotatebox[origin = lb]{45}{\textbf{\scriptsize #1}}}}}

\usepackage{pdfrender}

\usepackage{graphicx}
\makeatletter
\def\maxwidth#1{\ifdim\Gin@nat@width>#1 #1\else\Gin@nat@width\fi}
\makeatother

\usepackage{contour}
\contourlength{1pt} %how thick each copy is
\contournumber{20}  %number of copies

\usepackage{pifont}

\def\ccheck{\color{black}{\ding{52}}}
\def\ccross{\color{black}{\ding{56}}}

\title{\model \space v1.2: A more efficient family of OlmoEarth models}

\authorOne[\starOlmo]{Team OlmoEarth}
\authorTwo[\coreContrib]{Gabriel Tseng}
\authorTwo[\coreContrib]{Yawen Zhang}
\authorTwo[\coreContrib]{Favyen Bastani}
\authorTwo[\coreContrib]{Henry Herzog}
\authorTwo[\coreContrib]{Joseph Redmon}
\authorThree[\coreContrib]{Hadrien Sablon}
\authorThree[\coreContrib]{Piper Wolters}
\authorThree[\coreContrib]{Ando Shah}
\authorThree{Patrick Alan Johnson}
\authorFour{Christopher Wilhelm}
\authorFour[\coreContrib]{Patrick Beukema}

\affiliation{Allen Institute for AI}
\contribution[]{\starOlmo {\model} was a team effort.\\
\coreContrib{Equal contribution from modeling team}.}

\abstract{We present a set of improvements to the \model \space family. These improvements allow us to cut compute costs during training ($3.0 \times$ reduction in GPU hours required to train our Base models) and inference ($5.3\times$ reductions in MACs on Sentinel-2 tasks), while maintaining the models' overall performance. All training code is available at {\modelurl}.

}

\metadata[\quad\aitoo Platform:]{\href{https://olmoearth.allenai.org/}{\texttt{olmoearth.allenai.org}}}

\metadata[\vspace{.5em}\quad\github Training Code:]{
    \href{https://github.com/allenai/olmoearth_pretrain}{\texttt{olmoearth\_pretrain}}
}

\metadata[\vspace{.5em}\quad\huggingface OlmoEarth Pre-trained Models:]{
    \href{https://huggingface.co/allenai/OlmoEarth-v1_2-Nano}{\texttt{OlmoEarth-v1\_2-Nano}} \quad
    \href{https://huggingface.co/allenai/OlmoEarth-v1_2-Tiny}{\texttt{OlmoEarth-v1\_2-Tiny}}

    \hspace{17.4em}
    \href{https://huggingface.co/allenai/OlmoEarth-v1_2-Small}{\texttt{OlmoEarth-v1\_2-Small}} \quad
    \href{https://huggingface.co/allenai/OlmoEarth-v1_2-Base}{\texttt{OlmoEarth-v1\_2-Base}}
}

\metadata[\vspace{.5em}\quad\huggingface OlmoEarth Pre-training Dataset:]{
\href{https://huggingface.co/datasets/allenai/olmoearth_pretrain_dataset}{\texttt{olmoearth\_pretrain\_dataset}}}
\metadata[\vspace{.5em}\quad\emailLogo Contact:]{\color{ai2accent}{\texttt{olmoearth@allenai.org}}}

\begin{document}

\definecolor{colordefault}{HTML}{\colordefault}
\definecolor{colorgrey}{HTML}{\colorgrey}
\definecolor{colorone}{HTML}{\colorone}
\definecolor{colortwo}{HTML}{\colortwo}
\definecolor{colorthree}{HTML}{\colorthree}
\definecolor{colorfour}{HTML}{\colorfour}
\definecolor{colorfive}{HTML}{\colorfive}
\definecolor{colorsix}{HTML}{\colorsix}
\maketitle
% \newpage
%\input{sec/1_intro_taylors_version}
% \input{sec/1_intro}
\section{Introduction}
\label{sec:intro}

OlmoEarth is a family of Earth observation foundation models, which obtains state-of-the-art results across a range of tasks \cite{olmoearth}. In this technical report, we discuss a number of changes which (1) reduce computational costs at training and inference time while (2) maintaining (and in some cases improving) performance on downstream tasks. We combine these changes into OlmoEarth v1.2 \footnote{For clarity, we combine the changes from both OlmoEarth v1.1 and v1.2 in this report. A report which isolates the v1.1 changes is available at \url{https://arxiv.org/abs/2605.20804v1}.} (throughout this report, we will refer to the original OlmoEarth models - as described in \citet{olmoearth} - as ``OlmoEarth v1''). 

In Section \ref{sec:model}, we describe the changes between the v1 and v1.2 models. 

In Section \ref{sec:experiments}, we present the experimental results of OlmoEarth v1.2. In general we suggest using OlmoEarth v1.2 as a drop-in replacement for OlmoEarth v1.

  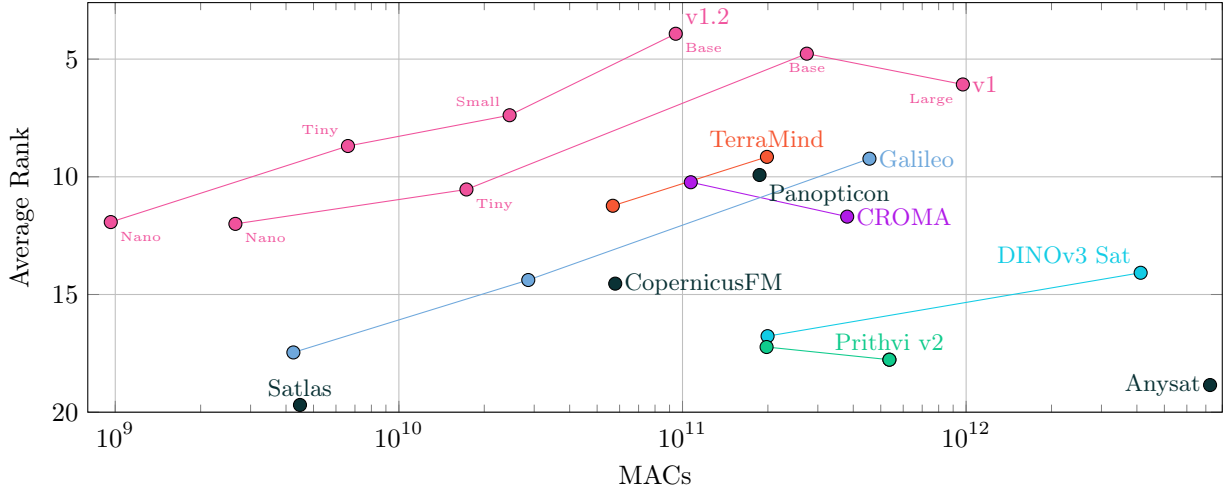
\begin{figure}[t]
      \centering
      \begin{tikzpicture}
          % Marker style — defined once
          \tikzset{modelmarker/.style={circle, inner sep=1.73329pt, draw=black}}

          \begin{axis}[
              xmode=log,
              width=\columnwidth,
              height=7cm,
              xmin=3e9, xmax=8e12,
              ymin=2.6, ymax=21,
              xtick={1e9, 1e10, 1e11, 1e12, 1e13},
              xlabel={MACs},
              ylabel={Average Rank},
              grid=major,
              y dir=reverse
          ]

            % ============================================================
            % MODEL DATA — MACs corrected to include attention matmuls
            % (thop counts only Linear/Conv; QK^T and AV were uncounted).
            % Ranks are unaffected. Factor vs published in each comment.
            % ============================================================
            % CROMA
            \coordinate (croma-1)     at (axis cs:111924196549,  10.23076923);  % 1.05x
            \coordinate (croma-2)     at (axis cs:394294330762,  11.69230769);  % 1.04x
            % DINOv3 Sat
            \coordinate (dinov3-1)    at (axis cs:276638507638,  16.76923077);  % 1.39x
            \coordinate (dinov3-2)    at (axis cs:5963011946023, 14.07692308);  % 1.45x
            % Galileo
            \coordinate (galileo-1)   at (axis cs:92241634658,   17.46153846);  % 21.72x
            \coordinate (galileo-2)   at (axis cs:424512375434,  14.38461538);  % 14.85x
            \coordinate (galileo-3)   at (axis cs:2039829824197, 9.230769231);  % 4.47x
            % Prithvi v2
            \coordinate (prithvi-1)   at (axis cs:261633664866,  17.23076923);  % 1.32x
            \coordinate (prithvi-2)   at (axis cs:717538667914,  17.76923077);  % 1.34x
            % TerraMind
            \coordinate (terramind-1) at (axis cs:59136953738,   11.23076923);  % 1.04x
            \coordinate (terramind-2) at (axis cs:204739722004,  9.153846154);  % 1.03x
            % OlmoEarth v1
            \coordinate (v1-1) at (axis cs:34316909647,   12);            % 12.94x
            \coordinate (v1-2) at (axis cs:159805931993,  10.53846154);   % 9.23x
            \coordinate (v1-3) at (axis cs:844292363815,  4.769230769);   % 3.08x
            \coordinate (v1-4) at (axis cs:2493645662050, 6.076923077);   % 2.56x
            % OlmoEarth v1.2
            \coordinate (v12-1) at (axis cs:4479262247,   11.92307692);   % 4.65x
            \coordinate (v12-2) at (axis cs:22432871188,  8.692307692);   % 3.39x
            \coordinate (v12-3) at (axis cs:56205749642,  7.384615385);   % 2.29x
            \coordinate (v12-4) at (axis cs:157929759980, 3.923076923);   % 1.67x
            % Single-point models
            \coordinate (anysat)       at (axis cs:643495127188, 18.84615385);  % 0.12x (published inflated by a thop bug)
            \coordinate (clay)         at (axis cs:270365348805, 20.53846154);  % 1.04x
            \coordinate (copernicusfm) at (axis cs:61847048921,  14.53846154);  % 1.07x
            \coordinate (panopticon)   at (axis cs:190944609674, 9.923076923);  % 1.02x
            \coordinate (satlas)       at (axis cs:15906683353,  19.69230769);  % 3.55x
        
          % ============================================================
          % Lines through model families
          % ============================================================
          \draw[colortwo,-]   (croma-1) -- (croma-2);
          \draw[colorfive,-]  (dinov3-1) -- (dinov3-2);
          \draw[colorthree,-] (galileo-1) -- (galileo-2) -- (galileo-3);
          \draw[colorfour,-]  (prithvi-1) -- (prithvi-2);
          \draw[colorsix,-]   (terramind-1) -- (terramind-2);
          \draw[colorone,-]   (v1-1) -- (v1-2) -- (v1-3) -- (v1-4);
          \draw[colorone,-]   (v12-1) -- (v12-2) -- (v12-3) -- (v12-4);

          % ============================================================
          % Single-point models
          % ============================================================
          \node[modelmarker, fill=colordefault] at (anysat) {};
          \node[anchor=east,  text=colordefault] at (anysat)       {\small Anysat };
          \node[modelmarker, fill=colordefault] at (clay) {};
          \node[anchor=south, text=colordefault] at (clay)         {\small Clay };
          \node[modelmarker, fill=colordefault] at (copernicusfm) {};
          \node[anchor=west, text=colordefault] at (copernicusfm) {\small CopernicusFM };
          \node[modelmarker, fill=colordefault] at (panopticon) {};
          \node[anchor=north west, text=colordefault] at (panopticon)   {\small Panopticon };
          \node[modelmarker, fill=colordefault] at (satlas) {};
          \node[anchor=south, text=colordefault] at (satlas)       {\small Satlas };

          % CROMA
          \node[modelmarker, fill=colortwo] at (croma-1) {};
          \node[modelmarker, fill=colortwo] at (croma-2) {};
          \node[anchor=west, text=colortwo] at (croma-2) {\small CROMA };

          % DINOv3 Sat
          \node[modelmarker, fill=colorfive] at (dinov3-1) {};
          \node[modelmarker, fill=colorfive] at (dinov3-2) {};
          \node[anchor=south east, text=colorfive] at (dinov3-2) {\small DINOv3 Sat };

          % Galileo
          \node[modelmarker, fill=colorthree] at (galileo-1) {};
          \node[modelmarker, fill=colorthree] at (galileo-2) {};
          \node[modelmarker, fill=colorthree] at (galileo-3) {};
          \node[anchor=west, text=colorthree] at (galileo-3) {\small Galileo };

          % Prithvi v2
          \node[modelmarker, fill=colorfour] at (prithvi-1) {};
          \node[modelmarker, fill=colorfour] at (prithvi-2) {};
          \node[modelmarker, fill=colorfour] at (prithvi-2) {};
          \node[anchor=south, text=colorfour] at (prithvi-2) {\small Prithvi v2 };

          % TerraMind
          \node[modelmarker, fill=colorsix] at (terramind-1) {};
          \node[modelmarker, fill=colorsix] at (terramind-2) {};
          \node[anchor=south east, text=colorsix] at (terramind-2) {\small TerraMind };

          % OlmoEarth v1
          \node[modelmarker, fill=colorone] at (v1-1) {};
          \node[modelmarker, fill=colorone] at (v1-2) {};
          \node[modelmarker, fill=colorone] at (v1-3) {};
          \node[modelmarker, fill=colorone] at (v1-4) {};
          \node[anchor=west, text=colorone] at (v1-4) {\small v1};

          % OlmoEarth v1.1
          \node[modelmarker, fill=colorone] at (v12-1) {};
          \node[modelmarker, fill=colorone] at (v12-2) {};
          \node[modelmarker, fill=colorone] at (v12-3) {};
          \node[modelmarker, fill=colorone] at (v12-4) {};
          \node[anchor=south west, text=colorone] at (v12-4) {\small v1.2};

          % add node markers to make the size explicit, for v1.1
          \node[anchor=north west, text=colorone] at (v12-1) {\tiny Nano };
          \node[anchor=south east, text=colorone] at (v12-2) {\tiny Tiny };
          \node[anchor=south east, text=colorone] at (v12-3) {\tiny Small };
          \node[anchor=north west, text=colorone] at (v12-4) {\tiny Base };

          % and for v1
          \node[anchor=north west, text=colorone] at (v1-1) {\tiny Nano };
          \node[anchor=north, text=colorone] at (v1-2) {\tiny Tiny };
          \node[anchor=north, text=colorone] at (v1-3) {\tiny Base };
          \node[anchor=north east, text=colorone] at (v1-4) {\tiny Large};

          \end{axis}
      \end{tikzpicture}
      \caption{OlmoEarth v1.2 defines a Pareto frontier of
  performance vs. computational efficiency averaged across 13 embedding
  tasks (measured by kNN and linear probing)\protect\footnotemark. The chart
  shows average Multiply-Accumulate operations (MACs) to encode one example across
  all tasks (input size varies by task) - each line connects model sizes within a family. OlmoEarth v1.2 Base requires $5.3 \times$ fewer MACs than OlmoEarth v1 Base. For easier comparison between the v1 and v1.2 families, we note their model sizes. See Table
  \ref{tab:knnlp} for full
  results.}
      \label{fig:result_summary}
  \end{figure}

\section{OlmoEarth v1.2}
\label{sec:model}

The OlmoEarth models (both v1 and v1.2) are encoder-decoder vision-transformer models trained via masked image modeling. The models are trained on multimodal, multitemporal remote sensing inputs. During training, part of the input is masked, and the model learns to reconstruct the masked part based on the unmasked portion. 

OlmoEarth v1.2 has the same architecture and training data as OlmoEarth v1. These are discussed in detail in Sections 2.1 and 2.2 of \citet{olmoearth}. We improve the v1 models along three dimensions, and present these changes in the sections below.

\begin{itemize}
\item \textbf{Efficiency}: Remote sensing models are run at scale, so the computational costs of these models are a primary consideration in their adoption \cite{worldcerealsbenchmark}. v1 splits modalities into multiple band-sets when constructing tokens; in Section \ref{sec:single_bandsets} we use a single band-set per modality, reducing the computational footprint of v1.2 by $3\times$ compared to v1.  
\item \textbf{Performance}: We introduce two changes which improve the performance (i.e. accuracy) of the models: we revisit v1's loss function in Section \ref{sec:new_loss} and we revisit v1's masking function in Section \ref{sec:masking}.
\item \textbf{Reducing artifacts}: We observe that v1 introduces striping artifacts in its embeddings \footnote{\url{https://github.com/allenai/olmoearth_pretrain/issues/499}}. 
In Section \ref{sec:rope}, we introduce rotary positional encodings (RoPE), which removes these artifacts.
\end{itemize}

Smaller changes are discussed in Section \ref{sec:pretraining}. We begin below by recapping OlmoEarth v1 in Section \ref{sec:v1}.

\footnotetext{We rank models over all tasks every model can perform and average the ranks. Specifically these tasks are the Sentinel-2 versions of: m-bigearthnet, m-so2sat, m-brick-kiln, m-eurosat, BreizhCrops, CropHarvest-Togo, CropHarvest-PRC, m-cashewplant, m-SA-crop-type, PASTIS, MADOS, AWF, Nandi.}

\subsection{Background: OlmoEarth v1} \label{sec:v1}

We train OlmoEarth v1 using ``Latent Masked Image Modeling of Linear, Invariant Token Embeddings'' (\textbf{LatentMIM Lite}). In this setup, the raw data are transformed into tokens and the model is trained with a contrastive loss in this token space. 

To get the target tokens, we first patchify the raw data and then pass the patches through a frozen copy of the encoder's projection layer. As a result, the targets are just linear projections of the input patches (hence ``Linear, Invariant Token Embeddings''). In addition to the token-level contrastive loss, we also use an instance-level contrastive loss. For each batch, we perform two forward passes with different masks applied. We then mean-pool the tokens for each instance and apply an InfoNCE loss, where positives are two masked views of the same instance and negatives are different instances.

We train OlmoEarth v1 on three satellite modalities and six derived maps:

\begin{table}[h]
\centering
    \begin{tabular}{l|ll}
    Observations & Maps & \\
    \hline
    Sentinel-1 & WorldCereal \cite{van2023worldcereal} & OpenStreetMap \cite{OpenStreetMap} \\
    Sentinel-2 & WorldCover \cite{zanaga2022esa} & Cropland Data Layer \cite{USDA_NASS_Cropland_Data_Layer}\\
    Landsat-8 & SRTM \cite{SRTM} & Canopy Height Map \cite{tolan2024very}\\
    \end{tabular}
\end{table}

Observations are seen by the encoder, and can also be targets. Maps are only used as targets.

OlmoEarth v1.2 has an identical training dataset to v1, and follows the same basic formula for training. 

\subsection{Single bandsets per modality} \label{sec:single_bandsets}

OlmoEarth v1 splits Sentinel-2 and Landsat into ``bandsets'' - collections of bands which are tokenized together. Bandsets group bands by resolution - for instance, the Sentinel-2 bands are split into 3 band sets for the 10m, 20m and 60m resolution bands. This means a Sentinel-2 input is tokenized into $H_p \times W_p \times T \times 3$ tokens, where $H_p$ and $W_p$ are the spatial patch grid dimensions, $T$ is the number of timesteps, and $3$ is the number of bandsets. OlmoEarth v1.2 groups all bands in a modality into a single bandset, yielding $3 \times$ fewer tokens for a Sentinel-2 input ($H_p \times W_p \times T \times 1$ tokens). \textbf{This leads to a $\approx 5 \times$ reduction in MACs required to encode our evaluation inputs.} 

However, naively collapsing all bands into a single token can lead to significant performance reductions in certain tasks: we observed the m-eurosat kNN score drop from ${\sim}94 \%$ to ${\sim}84 \%$ when consolidating the 3 Sentinel-2 bandsets into a single bandset (Table \ref{tab:ablations}). When modalities are split into multiple band-sets, the model has to reconstruct some band-sets given others within the same modality. We hypothesized this provided an important learning signal which was lost when naively aggregating band-sets from a modality. Recovering performance required the following two changes: (1) random band dropout, and (2) a non-linear projection layer.

\begin{wrapfigure}{l}{0.4\textwidth}
    \centering
    \includegraphics[width=0.4\textwidth]{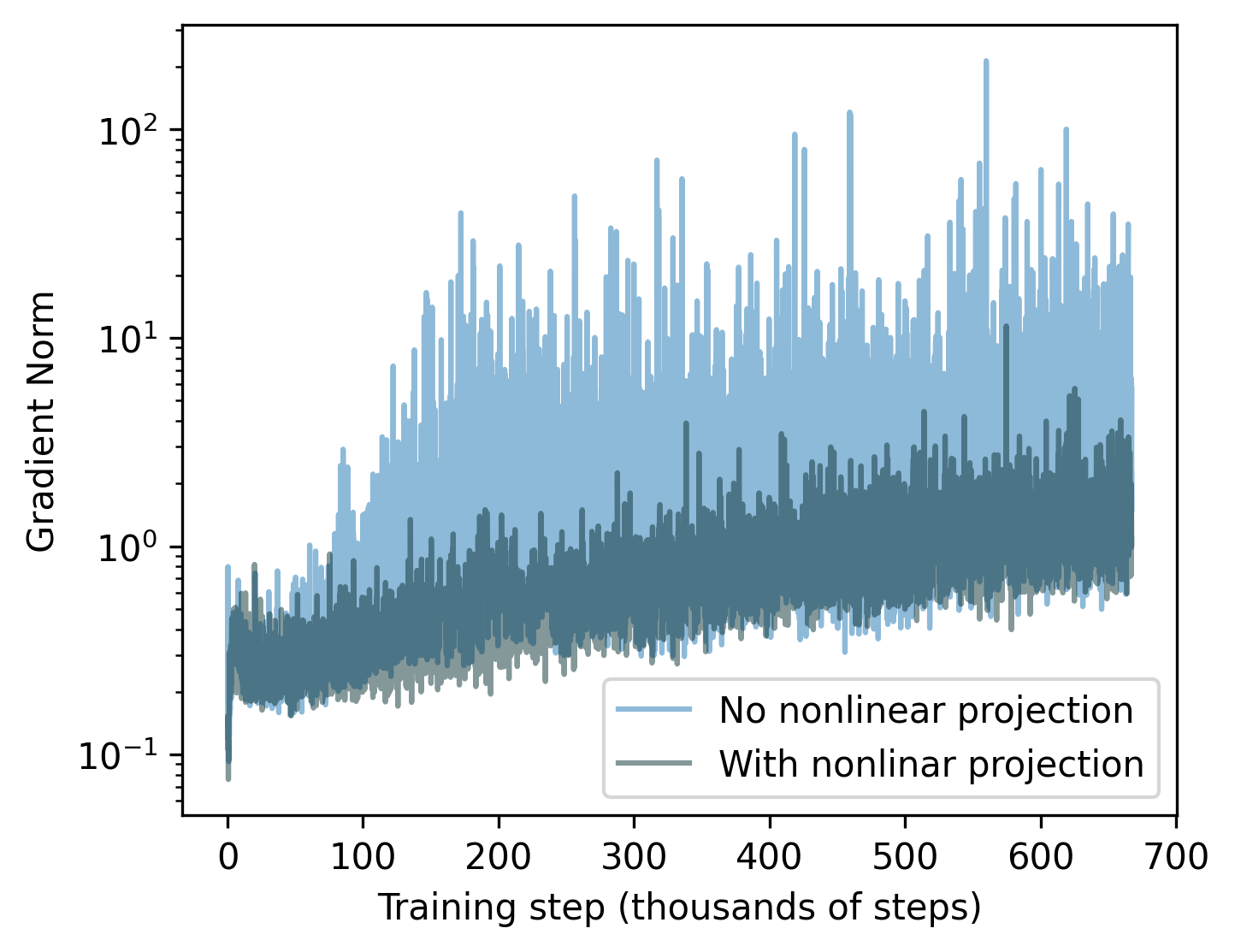}
    \caption{Gradient norms for Base models during pre-training with and without the nonlinear projection. We apply a log-scale to the y-axis.} \label{fig:norms}
\end{wrapfigure}

\paragraph{Random band dropout} To encourage the model to learn cross-band relationships within each modality, we introduced \textbf{random band dropout}. During training, each band is independently zeroed out with probability $r$ in the online encoder's input, where $r$ is uniformly sampled from $[0, r_{\max} = 0.2]$ per forward pass, while the target encoder always sees the full set of bands. We apply random band dropout only to Sentinel-2 (12 bands) and Landsat (11 bands), and exclude Sentinel-1 since it only has 2 bands. Since the training objective requires the prediction to match the target encoder's output computed from all bands, the model is forced to infer missing band information from the remaining bands, yielding richer cross-band representations within each token. This approach largely recovers the m-eurosat kNN performance to match the multi-bandset baseline.

\paragraph{A non-linear projection layer} While random band dropout recovered the performance of the frozen model (i.e. under a $k$NN or linear probing regime), we observed extremely high gradient norms during training. This impacted model stability during pretraining: several runs would experience an increasing loss, and a degradation on our evaluation metrics. In addition, finetuning performance of the models decreased; we hypothesize that the high gradient norms made the models harder to finetune. Adding a \textit{nonlinear} projection layer resolved these gradient issues and improved finetuning performance. Previously, we would linearly project pixels to token space: for a patch size $P$, we would construct tokens by dividing the image into $\frac{H}{P}, \frac{W}{P}$ sections and (for each section) computing $f_W : \mathbb{R}^{P\times P\times C} \to \mathbb{R}^{D}$. We now first map the pixels to a hidden dimension $H$, $f_{W_H} : \mathbb{R}^{P\times P\times C} \to \mathbb{R}^{P\times P\times H}$ (where $f_{W_H}$ includes a ReLU nonlinearity). We then map these to tokens, $f_W : \mathbb{R}^{P\times P\times H} \to \mathbb{R}^{D}$. We use $H = 64$ for Base and Tiny, and $H = 12$ for Nano. Figure \ref{fig:norms} shows the significant impact this has on gradient norms during pre-training. \citet{xiao2021early} had previously investigated nonlinear projections in ViTs, and found they improve optimization stability when training on natural images.

\subsection{An updated masking function} \label{sec:masking}

OlmoEarth v1's masking strategy works in two stages: first, random masking is applied on the whole input, i.e., assigning tokens to either be seen by the encoder (unmasked) or used to construct target tokens (masked). Second, bandset-level masking is applied - bandsets are selected to be in specific states (e.g. if a bandset is set to be masked, all unmasked tokens would be ignored so that they would be omitted from the loss entirely). A consequence of this is that tokens which might otherwise provide a learning signal would get set to ``ignore''. This issue is especially acute for the ``target-only'' modalities (maps, which are only ever used as targets). Since we first apply random masking (masking ratio = 0.5) to maps and then suppress all the unmasked tokens, we effectively remove 50\% of all map tokens from the learning signal. 

\sethlcolor{colorone!25}
\begin{table*}[htpb]
    \centering
    \resizebox{\textwidth}{!}{%

% Slightly bigger checkmarks here compared to the ablation study table.
    
    \begin{tabular}{ll|cccccccccccccccccc|cccccc}

\hdr{} & \hdr{} & \hdr{m-bigearthnet} & \hdr{m-so2sat} & \hdr{m-brick-kiln} & \hdr{m-forestnet} & \hdr{m-eurosat} & \hdr{BreizhCrops} & \hdr{CropHarvest-PRC} & \hdr{CropHarvest-PRC} & \hdr{CropHarvest-PRC} & \hdr{CropHarvest-Togo} & \hdr{CropHarvest-Togo} & \hdr{CropHarvest-Togo} & \hdr{m-cashewplant} & \hdr{m-SA-crop-type} & \hdr{PASTIS} & \hdr{PASTIS} & \hdr{MADOS} & \hdr{Sen1Floods11} & \hdr{AWF} & \hdr{AWF} & \hdr{AWF} & \hdr{Nandi} & \hdr{Nandi} & \hdr{Nandi} \\
 & Modalities & S2 & S2 & S2 & L8 & S2 & S2 & S1 & S2 & S1,S2 & S1 & S2 & S1,S2 & S2 & S2 & S1 & S2 & S2 & S1 & L8 & S1 & S2 & L8 & S1 & S2 \\
 & Time series & \ccross & \ccross & \ccross & \ccross & \ccross & \ccheck & \ccheck & \ccheck & \ccheck & \ccheck & \ccheck & \ccheck & \ccross & \ccross & \ccheck & \ccheck & \ccross & \ccross & \ccheck & \ccheck & \ccheck & \ccheck & \ccheck & \ccheck \\
 & Method & kNN & kNN & kNN & kNN & kNN & LP & LP & LP & LP & LP & LP & LP & LP & LP & LP & LP & LP & LP & kNN & kNN & kNN & kNN & kNN & kNN \\
Model & Metric & $\mu \text{F1}$ & Acc. & Acc. & Acc. & Acc. & Acc. & Acc. & Acc. & Acc. & Acc. & Acc. & Acc. & mIOU & mIOU & mIOU & mIOU & mIOU & mIOU & Acc. & Acc. & Acc. & Acc. & Acc. & Acc. \\
\hline
Anysat & ViT Base & 54.5 & 36.5 & 84.5 & 34.0 & 80.4 & 62.7 & 57.0 & 74.1 & 72.3 & 69.6 & 78.4 & 74.5 & 24.6 & 27.2 & 24.2 & 41.9 & 41.3 & 77.8 & 60.0 & 60.0 & 64.0 & 47.4 & 20.5 & 55.8 \\
Clay & ViT Large & 48.8 & 38.7 & 90.8 & 40.3 & 86.3 & 57.0 & 56.7 & 66.5 & 63.4 & 78.4 & 67.0 & 67.3 & 30.8 & 23.1 & 19.9 & 22.6 & 47.5 & 78.9 & 59.5 & 61.5 & 56.5 & 52.2 & 23.4 & 57.0 \\
CopernicusFM & ViT Base & 64.6 & 50.3 & 85.9 & - & 84.7 & 65.5 & 55.1 & 72.8 & 74.4 & 77.5 & 75.8 & 70.6 & 32.2 & 28.4 & 15.9 & 32.1 & 63.9 & 77.6 & - & 59.0 & 67.0 & - & 24.4 & 59.8 \\
CROMA & ViT Base & 61.3 & 51.3 & 92.0 & - & 84.6 & 69.0 & 56.8 & 74.3 & 75.1 & 76.8 & 80.7 & 76.5 & 24.9 & 30.3 & 26.3 & 44.7 & 60.4 & 78.8 & - & 67.5 & \textbf{79.0} & - & 24.6 & 68.2 \\
CROMA & ViT Large & 59.2 & 48.2 & 91.7 & - & 85.8 & 68.2 & 56.2 & 72.7 & 71.0 & 79.4 & 76.5 & 81.0 & 27.0 & 30.4 & 25.9 & 42.7 & 66.4 & 78.8 & - & 62.5 & 71.0 & - & 26.1 & 68.2 \\
DINOv3 & ViT Base & 51.0 & 47.1 & 91.3 & 43.3 & 86.6 & 31.3 & - & 64.5 & - & - & 67.0 & - & 23.5 & 26.7 & - & 18.1 & 53.5 & - & 48.5 & - & 61.0 & 42.5 & - & 54.1 \\
DINOv3 & ViT Large & 55.8 & 45.3 & 89.9 & 46.0 & 84.0 & 31.3 & - & 66.1 & - & - & 68.3 & - & 24.5 & 26.1 & - & 17.4 & 52.4 & - & 43.5 & - & 60.5 & 39.9 & - & 49.8 \\
DINOv3 & ViT Huge+ & 57.0 & 45.7 & 88.2 & 46.9 & 86.1 & 31.3 & - & 68.7 & - & - & 68.3 & - & 25.1 & 26.7 & - & 17.4 & 48.1 & - & 47.5 & - & 54.0 & 43.5 & - & 55.4 \\
DINOv3 & ViT 7B & 60.8 & 46.6 & 91.3 & 48.0 & 85.2 & 31.3 & - & 68.7 & - & - & 70.9 & - & 34.3 & 27.9 & - & 21.1 & 52.2 & - & 47.5 & - & 57.0 & 44.7 & - & 56.7 \\
DINOv3 Sat & ViT Large & 60.2 & 44.0 & 91.4 & 44.2 & 89.2 & 31.3 & - & 70.1 & - & - & 68.6 & - & 32.4 & 28.5 & - & 22.5 & 57.5 & - & 42.5 & - & 69.5 & 35.5 & - & 48.4 \\
DINOv3 Sat & ViT 7B & 61.6 & 50.1 & 91.4 & 47.0 & 91.3 & 31.3 & - & 72.2 & - & - & 71.9 & - & \textbf{54.1} & 31.7 & - & 26.3 & 59.7 & - & 49.5 & - & 68.5 & 31.8 & - & 42.5 \\
Galileo & ViT Nano & 55.0 & 53.7 & 90.9 & - & 89.4 & 66.3 & 60.9 & 74.4 & 72.4 & 75.2 & 70.3 & 78.1 & 21.2 & 19.5 & 19.2 & 19.1 & 53.1 & 78.6 & - & 65.5 & 67.5 & - & 24.4 & 64.2 \\
Galileo & ViT Tiny & 55.8 & 53.1 & 87.5 & - & 89.1 & 66.7 & 55.7 & 80.3 & 79.3 & 69.9 & 78.8 & 77.1 & 23.6 & 21.5 & 23.4 & 27.7 & 61.1 & 78.6 & - & 65.5 & 71.0 & - & 25.0 & 66.7 \\
Galileo & ViT Base & 58.3 & 55.7 & 91.1 & - & 92.8 & 69.7 & 60.9 & \textbf{81.9} & 79.3 & 67.3 & 80.1 & 77.5 & 28.9 & 25.3 & 28.0 & 39.6 & \textbf{68.4} & 79.4 & - & 66.5 & 72.5 & - & 26.8 & 67.3 \\
Panopticon & ViT Base & 64.9 & 60.5 & 92.9 & \textbf{52.3} & 95.2 & 57.7 & 55.9 & 75.9 & 75.6 & 72.2 & 72.9 & 76.5 & 32.7 & 27.3 & 23.7 & 30.2 & 66.1 & 78.0 & 66.0 & 65.0 & 71.5 & 60.4 & 22.9 & 65.2 \\
Presto & ViT Nano & - & - & - & - & - & 60.9 & 59.3 & 74.3 & 76.6 & 78.4 & 81.4 & 72.5 & - & - & 16.3 & 28.2 & - & - & - & 60.5 & 53.5 & - & 25.2 & 57.6 \\
Prithvi v2 & ViT Large & 53.8 & 42.6 & 90.3 & 37.9 & 82.2 & 66.2 & - & 72.0 & - & - & 71.9 & - & 51.1 & 27.3 & - & 37.2 & 55.6 & - & 57.0 & - & 55.5 & 57.7 & - & 59.3 \\
Prithvi v2 & ViT Huge & 52.1 & 39.5 & 89.9 & 41.4 & 81.2 & 66.1 & - & 71.7 & - & - & 69.9 & - & 49.2 & 29.7 & - & 37.5 & 58.2 & - & 59.0 & - & 55.0 & 57.9 & - & 55.9 \\
Satlas & Swin Base & 52.3 & 44.5 & 86.1 & 36.9 & 82.2 & 64.6 & 57.4 & 71.3 & - & 76.8 & 75.8 & - & 30.6 & 24.0 & 10.5 & 14.4 & 30.2 & 72.9 & 57.5 & 52.0 & 62.0 & 61.5 & 25.1 & 60.2 \\
TerraMind & ViT Base & 63.9 & 46.7 & 91.9 & - & 85.6 & 66.4 & 57.0 & 75.1 & 75.6 & 74.2 & 74.8 & 77.5 & 46.0 & 30.4 & 22.7 & 40.9 & 66.0 & 78.7 & - & 66.0 & 69.5 & - & 24.7 & 64.2 \\
TerraMind & ViT Large & 63.9 & 47.4 & 92.2 & - & 90.0 & 68.2 & 56.2 & 74.7 & 72.0 & 75.2 & 77.8 & 75.2 & 50.4 & 31.2 & 22.3 & 41.3 & 67.5 & 78.4 & - & 62.0 & 67.0 & - & 23.3 & 65.3 \\
TESSERA &  & - & - & - & - & - & - & - & - & 72.2 & - & - & 81.0 & - & - & - & - & - & - & - & - & - & - & - & - \\
\hline
{\model} v1 & ViT Nano & \cellcolor{colorone!25} 59.5 & 54.3 &  \cellcolor{colorone!25} \textbf{96.2} & \cellcolor{colorone!25} 38.8 &  \cellcolor{colorone!25} 89.9 & 64.1 & 58.0 & \cellcolor{colorone!25} 79.5 & 74.3 & 73.5 & \cellcolor{colorone!25} 81.7 & \cellcolor{colorone!25} \textbf{83.7} & 25.5 & \cellcolor{colorone!25} 23.6 & 18.1 &  35.0 & 55.2 & 78.2 & \cellcolor{colorone!25} 73.0 &  61.5 & 69.5 & 57.7 & 24.8 & 67.4 \\
{\model} v1 & ViT Tiny & 59.4 & \cellcolor{colorone!25} 61.8 &  92.0 & \cellcolor{colorone!25} 40.5 &  \cellcolor{colorone!25} 91.6 & 64.0 & 58.3 & \cellcolor{colorone!25} 78.6 &  \cellcolor{colorone!25} \textbf{80.3} &  \cellcolor{colorone!25} 75.8 &  \cellcolor{colorone!25} 85.6 & \cellcolor{colorone!25} 82.4 & 24.7 & 23.2 & 21.4 & 40.1 & 58.6 & 78.5 & \cellcolor{colorone!25} 75.5 & 64.0 & \cellcolor{colorone!25} 76.0 & 60.7 & 24.7 & \cellcolor{colorone!25} 69.0 \\
{\model} v1 & ViT Base & 62.4 &  \cellcolor{colorone!25} 67.7 & 93.3 & 41.9 &  \cellcolor{colorone!25} 94.7 & 70.9 & 56.8 & 73.4 &  75.4 &  \cellcolor{colorone!25} \textbf{80.1} & \cellcolor{colorone!25} \textbf{87.3} &  \cellcolor{colorone!25} 82.0 & \cellcolor{colorone!25} 32.3 & 28.9 & 29.7 & 50.6 & \cellcolor{colorone!25} 67.2 & \cellcolor{colorone!25} 79.2 & \cellcolor{colorone!25} \textbf{77.0} & \cellcolor{colorone!25} \textbf{68.5} & \cellcolor{colorone!25} 77.5 & \cellcolor{colorone!25} \textbf{67.9} & 26.5 & \cellcolor{colorone!25} \textbf{74.7} \\
{\model} v1 & VIT Large & 62.0 & \textbf{68.2} & 93.4 & 41.6 & \textbf{96.3} & 70.7 & 56.6 & 74.1 & 76.1 & 67.6 & 78.1 & 79.7 & 30.9 & 28.5 & 30.6 & 51.8 & 66.4 & \textbf{79.8} & 76.0 & 66.5 & 73.0 & 66.4 & 26.2 & 73.6 \\
\hline
{\model} v1.2 & ViT Nano &  \cellcolor{colorone!25} 59.5 & \cellcolor{colorone!25} 55.3 & 92.8 &  38.5 & 87.4 & \cellcolor{colorone!25} 67.5 & \cellcolor{colorone!25} \textbf{63.7} & 77.5 & \cellcolor{colorone!25} 80.2 & \cellcolor{colorone!25} 73.9 & 79.4 & 82.7 & \cellcolor{colorone!25} 26.8 & 22.4 & \cellcolor{colorone!25} 22.5 & \cellcolor{colorone!25} 35.9 & \cellcolor{colorone!25} 58.1 & \cellcolor{colorone!25} 78.7 & 71.5 & \cellcolor{colorone!25} 64.0 & \cellcolor{colorone!25} 71.5 & \cellcolor{colorone!25} 66.9 & \cellcolor{colorone!25} 25.9 & \cellcolor{colorone!25} 71.0 \\
{\model} v1.2 & ViT Tiny & \cellcolor{colorone!25} 61.7 & 58.8 & \cellcolor{colorone!25} 92.5 & 39.9 & 88.4 & \cellcolor{colorone!25} 70.6 & \cellcolor{colorone!25} 59.2 & 77.6 & 80.0 & 74.5 & 82.0 & 79.4 & \cellcolor{colorone!25} 27.1 & \cellcolor{colorone!25} 24.4 & \cellcolor{colorone!25} 24.9 & \cellcolor{colorone!25} 45.3 & \cellcolor{colorone!25} 63.0 & \cellcolor{colorone!25} 78.7 & 74.0 & \cellcolor{colorone!25} 64.5 & \cellcolor{colorone!25} 76.0 & \cellcolor{colorone!25} \textbf{67.9} & \cellcolor{colorone!25}  28.4 & 67.3 \\
{\model} v1.2 & ViT Small & 64.8 & 65.7 & 90.5 & 44.4 & 90.4 & 79.1 & 59.0 & 79.1 & 78.1 & 73.5 & 83.0 & 80.4 & 28.5 & 26.7 & 28.3 & 50.7 & 61.3 & 79.3 & 77.0 & 69.0 & 74.5 & 67.2 & \textbf{29.7} & 73.2 \\
{\model} v1.2 & ViT Base & \cellcolor{colorone!25} \textbf{65.5} & 67.3 & \cellcolor{colorone!25} 93.7 & \cellcolor{colorone!25} 45.4 & 92.3 & \cellcolor{colorone!25} \textbf{73.3} & \cellcolor{colorone!25} 58.2 & \cellcolor{colorone!25} 76.7 & \cellcolor{colorone!25} 77.7 & 74.8 & 81.4 & 81.0 & 29.6 & \cellcolor{colorone!25} \textbf{32.0} & \cellcolor{colorone!25} \textbf{31.1} & \cellcolor{colorone!25} \textbf{54.2} & \cellcolor{colorone!25} 67.2 & 78.9 & 74.5 & 68.0 & 75.0 & 65.7 & \cellcolor{colorone!25} 28.1 & 73.9 \\
% \hline

\end{tabular}}
    \caption{kNN/Linear probe results on research benchmarks and real-world tasks from our partners. We run kNN on single time-step classification tasks and linear probing on all other tasks. We sweep across data normalization strategies, feature pooling, and learning rate (for linear probing) and report the test set result for the best validation set performance. Not all models can run on all tasks due to incompatible input modalities. We \hl{highlight} which same-sized models win between v1 and v1.2.}
    \label{tab:knnlp}
\end{table*}    

We make the following changes to the masking function:

\begin{itemize}
\item \textbf{All maps used as targets}. The mapping modalities are a form of supervision for the model - they provide labels with which the model can be trained. To fully leverage these information-dense modalities, we set all tokens in the map modalities as ``unmasked''. 
\item \textbf{Time masking}. To encourage OlmoEarth to reason in time, we introduce temporal masking. Under this setting, timesteps are either completely masked or unmasked. This pushes the model to learn to recover observations at one timestep given observations at other timesteps. For each instance we choose time masking with probability $p_t = 0.5$, and random masking otherwise. We find that the model is not very sensitive to the choice of $p_t$, but that some masking helps (Figure \ref{fig:masking_ablation}). 
\end{itemize}

Time masking builds on a number of prior works which find that ``structured masking'' -- or masking which takes into account the spatio-temporal nature of remote sensing data, especially along the temporal dimension -- can outperform or complement random masking \cite{cong2022satmae,tseng2023lightweight,tseng2025galileo,szwarcman2024prithvi}.

\subsection{An updated loss function} \label{sec:new_loss}

OlmoEarth v1 is trained on a ``Modality Patch Discrimination'' loss. This loss modifies LatentMIM's Patch Discrimination loss \cite{wei2024towards} by only contrasting targets from the same modality. The purpose of this modification is to remove easy negatives, since the model can trivially separate different modalities.

Just as easy negatives can be a problem, extremely \textit{difficult} negatives can also be problematic (a phenomenon previously investigated by \citet{wu2023understanding} and \citet{huynh2022boosting}). For example, some tokens may be labeled entirely as water by WorldCover; it will be impossible for the model to distinguish between such tokens. We therefore remove extremely difficult negatives from the loss as well. We do this by comparing the cosine similarity of the target tokens; for each target token, any negative tokens with a cosine similarity of $\geq 0.999$ to the target are discarded. 

We only apply this thresholding to decode-only modalities. We do this because the encode-decode modalities (Sentinel-1, Sentinel-2, Landsat) are less likely to have tokens with similarities above this threshold (at patch-size 8, 0.50 \% of Sentinel-2 tokens have similarity $\geq$ 0.999 and across all patch sizes, 1.20\% of Sentinel-2 tokens have similarity $\geq$ 0.999). For the encode-decode modalities, we therefore avoid running the similarity comparison for efficiency during training.

\subsection{RoPE embeddings} \label{sec:rope}

OlmoEarth v1 and v1.2 can be used both as fully finetuneable models, or as frozen embedding extractors. v1 communicates four properties of each token via encodings: (1) the spatial position of each token, (2) the (indexed) position in time of each token, (3) the month of each token (to reflect seasonality) and (4) the modality each token came from. All of these encodings are applied via absolute positional encodings (APEs), which add the encodings to the tokens. These APEs could leak through to the final embeddings, introducing visual artifacts and impacting downstream performance. 

v1.2 replaces v1's absolute positional encodings with Rotary Positional Encodings (RoPE) \cite{su2024roformer}. RoPE is applied to the key and query vectors within the attention mechanism (unlike APEs, which are added to the tokens at the beginning of the forward pass and then unchanged). Briefly: each dimension in the KQ vectors gets paired (to yield $\texttt{dim}/ 2$ pairs, where $\texttt{dim}$ is the per-head dimension) and each pair gets rotated along a 2D circle according to the token's position. 

\begin{figure}[h]
\centering
\begin{minipage}{0.3\linewidth}
    \centering
    \includegraphics[width=\linewidth]{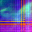}
    \caption*{APE}
\end{minipage}
\hspace{0.06\linewidth}
\begin{minipage}{0.3\linewidth}
    \centering
    \includegraphics[width=\linewidth]{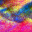}
    \caption*{RoPE}
\end{minipage}

\caption{PCA visualizations of OlmoEarth v1.2 embeddings over water, using APE or RoPE. APE produces grid-aligned artifacts under tiled inference, while RoPE reduces them.}
\label{fig:water-ape-rope}
\end{figure}

\begin{figure}[h]
\centering

\begin{minipage}{0.3\linewidth}
    \centering
    \includegraphics[width=\linewidth]{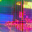}
    \caption*{APE}
\end{minipage}
\hspace{0.06\linewidth}
\begin{minipage}{0.3\linewidth}
    \centering
    \includegraphics[width=\linewidth]{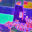}
    \caption*{RoPE}
\end{minipage}

\caption{PCA visualizations of OlmoEarth v1.2 embeddings on a tile from PASTIS dataset, using APE or RoPE. APE produces grid-aligned artifacts under tiled inference, while RoPE reduces them.}
\label{fig:pastis-rope}
\end{figure}

We use 3D-mixed RoPE \cite{heo2024rotary} - these encodings replace (i) 2D positional encodings and (ii) temporal encodings. Specifically, for every attention block (and every head within that attention block) we learn  $(\texttt{dim}/ 2) \times 3$ parameters (i.e. 3 parameters per dimension-pair): $\theta_x, \theta_y, \theta_t$. We rotate each dimension-pair according to the token's temporal position in the sequence $t$, and spatial position in the input $x,y$: $\texttt{angle} = \theta_t \times t + \theta_x \times x + \theta_y \times y$. The month encodings and modality encodings are unchanged from v1 (i.e. remain added to the tokens). This change significantly reduces visual artifacts (Figures \ref{fig:water-ape-rope} and \ref{fig:pastis-rope}). 

We ablate different RoPE implementation strategies in Appendix \ref{sec:app_rope}.

\subsection{Additional Changes} \label{sec:pretraining}

The single-bandset v1.2 models are less memory intensive than the v1 models. This allowed us to increase the micro-batch size to 64 (from 32), so that the instance-contrastive loss is now applied over 64 instances. We empirically find that reducing the weight on the instance contrastive loss from $0.1$ to $0.05$ improves performance.

In addition, we change our patch embedding from a convolution to a linear embedding. This linear layer increases throughput while being mathematically equivalent to the convolutional layer; we discuss this in more detail in Appendix \ref{sec:conv_to_linear}.
\section{Experiments \& Results}
\label{sec:experiments}

Aside from the modifications described above, we train OlmoEarth v1.2 identically to OlmoEarth v1. This training process is described in Section 3.1 of \citet{olmoearth}. To recap: we train OlmoEarth v1.2 using an AdamW optimizer with a batch size of $512$ for $667,200$ steps. We use a linear warmup of $8,000$ steps and then apply a cosine annealing learning rate for the remainder of training. 

We evaluate OlmoEarth v1.2 using the same suite of tests as for OlmoEarth v1. The OlmoEarth v1 evaluations benchmarked a wide range of pretrained models -- by reusing the evaluations, we can contextualize OlmoEarth v1.2's results against all these pretrained models. For this reason, all evaluation protocols are identical to those used in OlmoEarth v1 - while we summarize them below, full details are available in \citet{olmoearth}. 

OlmoEarth v1 and v1.2 were trained on identical training datasets, for the same number of steps. This makes comparisons between the two models especially meaningful, since differences in performance can be attributed to the algorithmic changes described above. However, the training cost of OlmoEarth v1.2 is significantly less than OlmoEarth v1; a v1 training run requires $2,989$ GPU hours, while a v1.2 training run requires $1,012$ GPU hours (a $3\times$ reduction). 

\subsection{kNN and Linear Probing Results}

We adopt the same $k$NN and linear probing regime as OlmoEarth v1. For all models, we sweep normalization strategies (pretraining vs. dataset statistics), pooling strategies (mean pooling vs. max pooling) and -- for linear probes -- learning rates. All results are in Table \ref{tab:knnlp}.

For $k$NN and linear probing, OlmoEarth v1.2 is generally competitive with OlmoEarth v1 and in some cases outperforms it. We do see some regressions: notably, v1.2 is worse than v1 on m-eurosat at all sizes, and is generally worse on the CropHarvest tasks. However, we do also see some significant improvements (for example on m-bigearthnet, BreizhCrops and PASTIS). In spite of OlmoEarth v1.2's significantly smaller computational footprint, performance changes between the models are small: when we average scores across all the tasks, OlmoEarth Nano goes from $60.1$ (v1) $\to 61.4$ (v1.2), OlmoEarth Tiny from $61.9 \to 62.8$ and OlmoEarth Base from $65.2 \to 65.2$.

\subsection{Finetuning Results} \label{sec:finetuning_results}

We adopt the same finetuning regime as OlmoEarth v1.
For the research tasks we apply a linear decoder and sweep learning rates for each model over $\{ 1 \times 10^{-4}, 5 \times 10^{-4}, 1 \times 10^{-3}\}$.
For the partner tasks, we apply task-specific decoders and finetune all models with the same learning rate ($10^{-4}$), except Nandi for which some models exhibit unstable learning and we sweep over $\{10^{-4}, 10^{-5}\}$. For all tasks, we freeze the encoder for the first 20\% of epochs, and then unfreeze it for the rest of training. We train research tasks for 50 epochs since they use a simpler head, and partner tasks for 100 epochs. For research tasks, we apply 1/10 the configured learning rate for the encoder parameters after unfreezing, while for partner tasks, we apply a uniform learning rate.
All results are in Table \ref{tab:finetune}.

\sethlcolor{colorone!25}
\begin{table*}[htpb]
    \centering
    \resizebox{\textwidth}{!}{%
    \begin{tabular}{ll|cccccccccc|ccccccccccccccccc}

\hdr{} & \hdr{} & \hdr{m-bigearthnet} & \hdr{m-so2sat} & \hdr{m-brick-kiln} & \hdr{m-forestnet} & \hdr{m-eurosat} & \hdr{m-cashewplant} & \hdr{m-SA-crop-type} & \hdr{PASTIS} & \hdr{MADOS} & \hdr{Sen1Floods11} & \hdr{AWF} & \hdr{AWF} & \hdr{GEA North Africa} & \hdr{Forest Loss Driver} & \hdr{Live Fuel Moisture Content} & \hdr{Live Fuel Moisture Content} & \hdr{Mangrove} & \hdr{Mangrove} & \hdr{Nandi} & \hdr{Nandi} & \hdr{Vessel Detection} & \hdr{Vessel Detection} & \hdr{Vessel Length} & \hdr{Vessel Type} & \hdr{Solar Farm Detection} \\
 & Modalities & S2 & S2 & S2 & L8 & S2 & S2 & S2 & S2 & S2 & S1 & S2 & S2, S1 & S2 & S2 & S2 & S2, S1 & S2 & S2, S1 & S2 & S2, S1 & L8 & S1 & S2 & S2 & S2 \\
 & Time series & \ccross & \ccross & \ccross & \ccross & \ccross & \ccross & \ccross & \ccheck & \ccross & \ccross & \ccheck & \ccheck & \ccheck & \ccheck & \ccheck & \ccheck & \ccheck & \ccheck & \ccheck & \ccheck & \ccross & \ccross & \ccross & \ccross & \ccheck \\
Model & Metric & $\mu \text{F1}$ & Acc. & Acc. & Acc. & Acc. & mIOU & mIOU & mIOU & mIOU & mIOU & Acc. & Acc. & Acc. & Acc. & L1 & L1 & Acc. & Acc. & Acc. & Acc. & F1 & F1 & L1 & Acc. & mIoU \\
\hline
Anysat & ViT Base & 68.4 & 56.7 & 98.7 & 51.6 & 95.9 & 80.4 & 34.2 & 60.9 & 63.3 & 77.4 & 78.0 & 83.0 & 59.3 & 84.6 & 19.1 & 19.4 & 96.9 & 97.1 & 78.4 & 76.7 & - & - & 73.4 & 43.5 & 82.6 \\
Clay & ViT Large & 65.7 & 61.3 & 98.7 & 49.2 & 95.8 & 73.9 & 33.4 & 48.9 & 68.9 & 78.5 & 77.0 & - & 53.4 & 93.3 & 24.8 & - & 96.6 & - & 30.9 & - & 70.7 & \textbf{79.9} & 16.4 & 68.3 & 82.2 \\
CopernicusFM & ViT Base & 71.3 & 66.8 & 98.1 & - & 98.5 & 78.7 & 33.6 & 54.6 & 66.0 & 78.6 & 79.0 & 79.0 & 58.8 & 90.0 & 25.2 & 24.5 & 97.1 & 97.1 & 68.2 & 55.9 & - & 77.4 & 16.7 & 69.6 & 77.6 \\
CROMA & ViT Base & 69.5 & 59.1 & 98.7 & - & 95.6 & 46.4 & 34.8 & 56.3 & 66.6 & 79.4 & 76.5 & 75.5 & 57.5 & 93.2 & 24.3 & 24.0 & 96.4 & 96.3 & 62.2 & 67.4 & - & - & 19.8 & 64.4 & 79.5 \\
CROMA & ViT Large & 71.8 & 58.9 & 97.8 & - & 97.5 & 47.8 & 36.0 & 58.1 & 68.8 & 79.4 & 79.5 & - & 56.6 & 92.3 & 24.6 & 24.1 & 96.6 & 96.2 & 76.4 & - & - & - & - & - & 81.7 \\
DINOv3 Sat & ViT Large & 69.9 & 63.3 & 98.9 & \textbf{59.2} & 96.7 & 80.6 & 34.5 & 42.8 & 64.7 & - & 34.5 & - & 43.0 & 80.4 & 90.2 & - & 65.8 & - & 35.8 & - & - & - & 30.3 & 54.8 & 70.7 \\
Galileo & ViT Base & 69.2 & 64.7 & 98.3 & - & 97.8 & 78.8 & 35.7 & 61.2 & 71.9 & 79.7 & 81.0 & 81.5 & \textbf{62.9} & 95.1 & 20.1 & 18.7 & 97.3 & 97.5 & \textbf{81.9} & 81.9 & - & 78.7 & 16.4 & 73.0 & 83.1 \\
Panopticon & ViT Base & 69.3 & 65.4 & \textbf{99.0} & 56.0 & 98.2 & 79.7 & 33.4 & 54.4 & 72.8 & 79.1 & 75.5 & 78.5 & 54.3 & 96.4 & 24.5 & 23.7 & 97.1 & 97.4 & 65.2 & 69.5 & 74.9 & 76.7 & 17.7 & 69.4 & 81.8 \\
Prithvi v2 & ViT Huge & 70.6 & 64.7 & 98.2 & - & 96.8 & 81.1 & 38.8 & 58.6 & 69.3 & - & 80.0 & - & 60.6 & 92.4 & - & - & 97.2 & - & 77.1 & - & 71.1 & - & 17.4 & 68.2 & 84.1 \\
Satlas & Swin Base & 72.7 & 65.1 & 98.7 & 56.0 & 97.0 & 77.0 & 37.8 & 57.4 & 60.5 & 78.5 & 78.0 & - & 56.1 & 63.3 & 25.0 & 24.6 & 96.6 & - & 47.6 & - & - & - & 16.2 & 71.6 & 83.3 \\
TerraMind & ViT Base & 72.6 & 66.1 & 98.5 & - & 97.6 & 80.9 & 39.2 & 59.9 & 73.2 & 79.5 & 84.0 & 82.0 & 49.8 & 96.4 & 24.3 & 23.8 & \textbf{97.7} & 96.8 & 66.1 & 79.3 & - & 79.6 & 18.1 & - & 83.5 \\
TerraMind & ViT Large & \textbf{74.0} & 65.4 & 98.1 & - & 97.8 & 81.3 & 41.1 & 60.9 & 71.5 & 79.5 & 81.5 & - & 51.1 & 93.9 & 24.5 & 24.5 & 96.5 & 96.9 & 66.1 & - & - & - & - & - & 83.0 \\
\hline
% {\model} (Random Init) & ViT Base & 61.0 & 48.9 & 94.7 & 41.7 & 80.3 & 43.0 & 27.5 & 43.9 & 45.6 & 77.0 & 62.5 & - & 52.9 & 52.7 & 20.9 & 20.5 & 96.3 & 96.4 & 60.6 & 56.1 & - & - & - & - & 74.1 \\
{\model} v1 & ViT Nano & 66.8 &  \cellcolor{colorone!25} 61.5 & 98.0 & 50.3 & \cellcolor{colorone!25} 95.3 & 39.5 & 35.4 & \cellcolor{colorone!25} 53.0 & \cellcolor{colorone!25} 60.6 & 78.8 & \cellcolor{colorone!25} 82.5 & \cellcolor{colorone!25} 82.5 & 61.1 & \cellcolor{colorone!25} 96.0 & \cellcolor{colorone!25} 20.4 & 19.7 & \cellcolor{colorone!25} 97.4 & \cellcolor{colorone!25} 97.4 & 75.6 & 74.8 & 70.2 & \cellcolor{colorone!25} 75.5 & 17.1 & 72.0 & \cellcolor{colorone!25} 82.1 \\
{\model} v1 & ViT Tiny & 69.6 & 63.5 & \cellcolor{colorone!25} 98.7 & 53.2 &  97.1 & 72.5 & 38.5 & 60.3 & \cellcolor{colorone!25} 71.5 & \cellcolor{colorone!25} 79.7 & \cellcolor{colorone!25} 85.0 & \cellcolor{colorone!25} 85.5 & \cellcolor{colorone!25} 60.6 & 97.7 & 19.8 & 19.2 & \cellcolor{colorone!25} 97.6 & \cellcolor{colorone!25} 97.7 & 78.2 & 76.4 & 74.4 & \cellcolor{colorone!25} 76.9 & 15.8 & 73.5 & 85.2 \\
{\model} v1 & ViT Base &  72.0 & 68.6 & 98.6 &  51.2 &  \cellcolor{colorone!25} \textbf{98.7} & 79.8 &  39.6 &  64.3 &  \cellcolor{colorone!25} 77.8 &  79.8 & \cellcolor{colorone!25} \textbf{87.0} & \cellcolor{colorone!25} \textbf{86.0} & \cellcolor{colorone!25} 62.4 & \cellcolor{colorone!25} 97.1 & \cellcolor{colorone!25} \textbf{18.5} & \cellcolor{colorone!25} \textbf{17.9} & \cellcolor{colorone!25} 97.6 & \cellcolor{colorone!25} \textbf{97.9} & \cellcolor{colorone!25} 81.8 & 82.2 & \cellcolor{colorone!25} 75.4 & 79.2 & 15.4 & 74.6 &  85.4 \\
{\model} v1 & ViT Large & 72.4 & 68.1 & 98.6 & 52.7 & 98.5 & 80.6 & 40.8 & \textbf{66.3} & \textbf{81.8} & 79.8 & 84.5 &  - & 58.8 & \textbf{97.9} & 19.9 & 18.5 & 97.6 & 97.6 & 81.0 &  - & - & - & - & - & 84.2 \\
\hline
{\model} v1.2 & ViT Nano & \cellcolor{colorone!25} 68.0 & 60.8 & \cellcolor{colorone!25} 98.5 & \cellcolor{colorone!25} 53.7 & \cellcolor{colorone!25} 95.3 & \cellcolor{colorone!25} 54.0 & \cellcolor{colorone!25} 36.3 & 52.7 & 58.1 & \cellcolor{colorone!25} 78.9 & 75 & 75.5 & \cellcolor{colorone!25} 61.5 & 95.4 & \cellcolor{colorone!25} 20.4 & 19.9 & 96.3 & 96.6 & \cellcolor{colorone!25} 79.2 & \cellcolor{colorone!25} 78.1 & \cellcolor{colorone!25} 71.1 & 75.1 & \cellcolor{colorone!25} 16.8 & \cellcolor{colorone!25} 72.4 & 80.1 \\
{\model} v1.2 & ViT Tiny & \cellcolor{colorone!25} 71.3 & \cellcolor{colorone!25} 67.4 & 98.4 & \cellcolor{colorone!25} 56.8 & \cellcolor{colorone!25} 97.3 & \cellcolor{colorone!25} 87.8 & \cellcolor{colorone!25} 41.7 & \cellcolor{colorone!25} 63.7 & 67.7 & 79.4 & 83.5 & 84.5 & 59.3 &  \cellcolor{colorone!25} 97.8 & \cellcolor{colorone!25} 19.5 & \cellcolor{colorone!25} 18.8 & 97.2 & 97.5 & \cellcolor{colorone!25} 80.7 & \cellcolor{colorone!25} 81.1 & \cellcolor{colorone!25} 75.2 & \cellcolor{colorone!25} 76.9 & \cellcolor{colorone!25} 14.3 & \cellcolor{colorone!25} 73.8 & \cellcolor{colorone!25} 86.6 \\
{\model} v1.2 & ViT Small & 72.0 & 66.5 & 98.7 & 56.5 & 97.6 & \textbf{95.8} & \textbf{44.0} & 64.7 & 69.7 & 79.8 & 83.0 & 82.0 & 62.0 & \textbf{97.9} & 19.4 & 19.0 & 97.5 & 97.7 & 78.3 & 80.2 & \textbf{75.6} & 81.9 & \textbf{14.2} & \textbf{75.9} & \textbf{87.5} \\
{\model} v1.2 & ViT Base & \cellcolor{colorone!25} 73.0 & \cellcolor{colorone!25} \textbf{70.8} & \cellcolor{colorone!25} 98.7 & \cellcolor{colorone!25} 56.1 & 98.0 & \cellcolor{colorone!25} 82.5 & \cellcolor{colorone!25} 43.5 & \cellcolor{colorone!25} 65.8 & 75.8 & \cellcolor{colorone!25} \textbf{80.5} & 83.0 & 85.0 & 60.2 & \cellcolor{colorone!25} 97.1 & 19.1 & 18.3 & 97.5 & 97.6 & 81.6 & \cellcolor{colorone!25} \textbf{82.5} & 74.8 & 78.3 & \cellcolor{colorone!25} 14.5 & \cellcolor{colorone!25} \textbf{75.9} & \cellcolor{colorone!25} 87.3 \\

\end{tabular}}
    \caption{Finetuning results on research benchmarks (left) and partner tasks (right). We train all models with the same recipe and report test set results for the model checkpoint with the best validation set performance. Some models are only compatible with a subset of tasks. Due to resource constraints, we do not fine-tune large models on all tasks. We \hl{highlight} which same-sized models win between v1 and v1.2.}
    \label{tab:finetune}
\end{table*}

Overall performance changes between the models are small. Finetuning performance on research tasks is a bit better, and finetuning performance on partner tasks is a bit worse. When averaged across all ``more is better'' tasks (i.e. excluding tasks where lower metrics indicate better performance), OlmoEarth Nano goes from $73.0$ (v1) $\to 73.3$ (v1.2), OlmoEarth Tiny goes from $77.0 \to 78.4$ and OlmoEarth Base goes from $79.0 \to 79.3$.

We investigate improvements to our finetuning recipe in Appendix \ref{app:llrd_ft}. 

\subsection{Ablations}

We ablate each of our changes in Table \ref{tab:ablations}, using models trained for $600,000$ steps. For these ablations, we evaluate our models with the validation sets of MADOS, m-eurosat, PASTIS and m-bigearthnet. Band dropout is critical to recover performance - without it, MADOS and m-eurosat performance drop significantly. The new masking strategy delivers small but consistent improvements on all tasks, and the new loss function improves all tasks except m-eurosat, where performance is similar. 

\sethlcolor{colorone!25}
{

\renewcommand*{\hdr}[1]{\textbf{#1}}

\begin{table}
    \centering
    % \resizebox{\textwidth}{!}{
    \scriptsize

    \begin{tabular}{ccccc|cccc}

Masking & $r_{max}$ & Projection & Loss & Encodings & MADOS & m-eurosat & PASTIS & m-bigearthnet \\
\hline
Updated & 0.2 & nonlinear & Masked negatives & RoPE & 73.6 & 91.8 & 55.9 & 65.0 \\
Updated & 0.2 & nonlinear & Masked negatives & \cellcolor{colorone!25} APE &  75.6 & 92.6 & 55.2 & 63.6  \\
\cellcolor{colorone!25} v1 & 0.2 & nonlinear & Masked negatives & \cellcolor{colorone!25} APE & 72.7 & 92.2 & 54.4 & 62.8 \\
Updated & \cellcolor{colorone!25} 0.0 & nonlinear &  Masked negatives & \cellcolor{colorone!25} APE & 69.3 & 84.1 & 54.8 & 63.9 \\
Updated & 0.2 & \cellcolor{colorone!25} linear &  Masked negatives & \cellcolor{colorone!25} APE & 75.5 & 93.9 & 54.7 &  62.0 \\
Updated & 0.2 & nonlinear & \cellcolor{colorone!25} v1 & \cellcolor{colorone!25} APE & 74.2 & 92.9 & 53.5 & 63.0 \\

\end{tabular}
%}
    \caption{Ablation experiments on the OlmoEarth v1.2 Base model, measuring validation performance with linear probing for MADOS (mIOU) \& PASTIS (mIOU), and $k$NN for m-eurosat (accuracy) \& m-bigearthnet ($\mu$F1). We \hl{highlight} the changes relative to v1.2. All experiments use single bandsets per modalities.} 
    \label{tab:ablations}
\end{table}
}

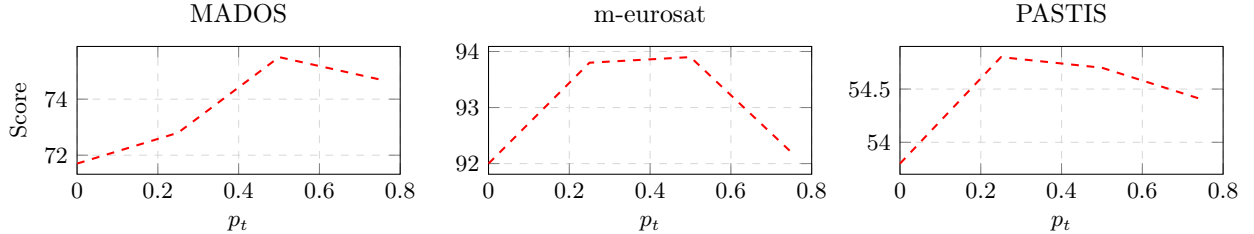
\begin{figure}[t]
      \centering
      \centering
    \resizebox{\textwidth}{!}{
      \begin{tikzpicture}
          \pgfplotsset{
              every axis/.style={
                  width=0.36\textwidth,
                  height=0.2\textwidth,
                  xlabel={$p_t$},
                  xmin=0, xmax=0.8,
                  xtick={0, 0.2, 0.4, 0.6, 0.8},
                  legend style={font=\small},
                  tick label style={font=\small},
                  label style={font=\small},
                  grid=major,
                  grid style={dashed, gray!30},
              },
          }

          \begin{axis}[
              name=plot1,
              ylabel={Score},
              title={MADOS},
          ]
        \addplot[mark=none, dashed, thick, red] coordinates {(0, 71.7) (0.25, 72.8) (0.5,75.5) (0.75, 74.7)};
          \end{axis}

          \begin{axis}[
              name=plot2,
              at={(plot1.east)},
              anchor=west,
              xshift=1.2cm,
              title={m-eurosat},
          ]
           \addplot[mark=none, dashed, thick, red] coordinates {(0, 92) (0.25, 93.8) (0.5, 93.9) (0.75, 92.2)};
          \end{axis}

          \begin{axis}[
              name=plot3,
              at={(plot2.east)},
              anchor=west,
              xshift=1.2cm,
              title={PASTIS},
          ]
          \addplot[mark=none, dashed, thick, red] coordinates {(0, 53.8) (0.25, 54.8) (0.5, 54.7) (0.75, 54.4)};
          \end{axis}

      \end{tikzpicture}}
      \caption{Model sensitivity to $p_t$, the ratio with which time masking is applied to the model. We measure validation performance on MADOS (mIOU), m-eurosat (accuracy) and PASTIS (mIOU) of $p_t \in \{0.0, 0.25, 0.5, 0.75\}$.}
      \label{fig:masking_ablation}
  \end{figure}

The addition of the time masking strategy introduces a new hyperparameter $p_t$, which defines how often we select time masking vs. random masking (see Section \ref{sec:masking} for more details). We measure model sensitivity to $p_t$ in Figure \ref{fig:masking_ablation} with models trained for $600,000$ steps. The model isn't very sensitive to this parameter; however, some amount of time masking ($p_t > 0$) is always helpful.

\subsection{Environmental Impact}

We use the same technique as OlmoEarth v1 to measure the pretraining cost of the OlmoEarth v1.2 models (Table \ref{tab:environmental-impact}). As with OlmoEarth v1, this estimate represents a lower bound since it does not account for hardware manufacturing, transportation, etc. Pretraining OlmoEarth v1.2 Base required 34\% of the GPU hours compared to pretraining v1 Base, and emitted 44\% of the tCO$_2$eq. 

\begin{table}[htpb]
    \centering
\resizebox{3.25in}{!}{%
\begin{tabular}{l | c c c c c c}
 & &  & & Energy & Carbon & Water \\
Model & Size & Hardware & GPU Hrs & (kWh) & (tCO$_2$eq) & (kL) \\
\hline
{\model} v1 & Nano & H100 & 1,149 & 195 & 0.08 & 0.30 \\
{\model} v1 & Tiny & H100 & 1,149 & 205 & 0.08 & 0.32 \\
{\model} v1 & Base & H100 & 2,989 & 803 & 0.32 & 1.24 \\
\hline
{\model} v1.2 & Nano & H100 & 1,151 & 191 & 0.09 & 0.30 \\
{\model} v1.2 & Tiny & H100 & 1,132 & 225 & 0.10 & 0.35 \\
{\model} v1.2 & Small & H100 & 1,253 & 265 & 0.12 & 0.41 \\
{\model} v1.2 & Base & H100 & 1,012 & 317 & 0.14 & 0.49 \\
\hline
\textbf{Total (v1.2)} & Overall & -- & 4,548 & 998.81 & 0.46 & 1.56 \\
\end{tabular}
}
\caption{Approximate environmental impact of pretraining {\model} v1 and v1.2.
}
\label{tab:environmental-impact}
\end{table}

\section{Discussion}

When making large scale maps, the costs of running and training the model dominate. When finetuning the model, 80\% of the compute cost consists of running finetuning (while 20\% consists of data export and preprocessing). When running inference at scale, 98\% of the cost is inference (while 2\% of the cost is data export \& preprocessing and map post-processing). OlmoEarth v1.2 therefore directly translates to cheaper mapping at scale; we hope this will make OlmoEarth more accessible to our partners and users of the OlmoEarth model and platform. 

We demonstrate that this reduction in computational cost does not need to penalize overall performance. We hope to continue finding the frontier of performance and cost. We observe that OlmoEarth v1.2 Tiny disproportionately benefits from the changes in this report; we are not sure why this is the case. One hypothesis is that the non-linear projection (Section \ref{sec:single_bandsets}) is relatively wider for the Tiny model (the projection width is 33\% of the model dimensionality) than for the Base (8.3 \%) and Nano (9.4 \%) models; having more capacity early in the model may be helpful. 

We release OlmoEarth v1.2 under the same ``OlmoEarth Artifact License'' as OlmoEarth v1. This open license restricts use for  military, defense-related, and extractive industry applications.

\clearpage
\bibliographystyle{ieeenat_fullname}
\bibliography{main}

\clearpage

\appendix

\section{RoPE encoding strategies} \label{sec:app_rope}

We tested four RoPE encoding strategies (where ``additive'' temporal represents APE, as in v1):

\begin{table}[h]
\centering
\begin{tabular}{llll}
\toprule
Variant & Spatial & Temporal & Frequencies \\
\midrule
\texttt{2D\_RoPE}            & axial 2D & additive  & fixed   \\
\texttt{2D\_RoPE\_mixed}     & mixed 2D & additive  & learned \\
\texttt{3D\_RoPE}        & axial 3D & rotary    & fixed   \\
\texttt{\detokenize{3D_RoPE_mixed}} & mixed 3D & rotary    & learned \\
\bottomrule
\end{tabular}
\end{table}

To evaluate these variants, we measured their linear probing and $k$NN performance on the \textbf{validation sets} of the evaluation tasks.

\sethlcolor{colorone!25}
\begin{table*}[htpb]
    \centering
    \resizebox{\textwidth}{!}{%
    \scriptsize

    \begin{tabular}{lcc|ccccccccccccccccc}

\hdr{} & \hdr{Avg.} & \hdr{Rank \ensuremath{\downarrow}} & \hdr{m-bigearthnet} & \hdr{m-so2sat} & \hdr{m-brick-kiln} & \hdr{m-forestnet} & \hdr{m-eurosat} & \hdr{m-cashewplant} & \hdr{m-SA-crop-type} & \hdr{PASTIS} & \hdr{PASTIS} & \hdr{MADOS} & \hdr{Sen1Floods11} & \hdr{AWF} & \hdr{AWF} & \hdr{AWF} & \hdr{Nandi} & \hdr{Nandi} & \hdr{Nandi} \\
 &  &  & S2 & S2 & S2 & L8 & S2 & S2 & S2 & S1 & S2 & S2 & S1 & L8 & S1 & S2 & L8 & S1 & S2 \\
 &  &  & \ccross & \ccross & \ccross & \ccross & \ccross & \ccross & \ccross & \ccheck & \ccheck & \ccross & \ccross & \ccheck & \ccheck & \ccheck & \ccheck & \ccheck & \ccheck \\
 &  &  & kNN & kNN & kNN & kNN & kNN & LP & LP & LP & LP & LP & LP & kNN & kNN & kNN & kNN & kNN & kNN \\
Variant & Avg. & Rank \ensuremath{\downarrow} & \ensuremath{\mu\mathrm{F1}} & Acc. & Acc. & Acc. & Acc. & mIOU & mIOU & mIOU & mIOU & mIOU & mIOU & Acc. & Acc. & Acc. & Acc. & Acc. & Acc. \\
\hline
APE & 61.2 & 3.91 & 63.6 & 64.2 & 95.7 & 45.9 & 92.5 & 32.1 & 32.9 & 31.1 & 55.2 & 75.3 & 79.7 & 68.4 & 67.9 & 70.5 & 65.6 & 30.4 & 68.9 \\

\texttt{2D\_RoPE} & 61.7 & 2.97 & 65.1 & 64.9 & 96.3 & 45.3 & 91.9 & 34.2 & 32.5 & 31.7 & 56.8 & 75.9 & 80.1 & 66.8 & 69.4 & 69.9 & 65.2 & 31.1 & 71.2 \\

\texttt{2D\_RoPE\_mixed} & \textbf{62.0} & 2.65 & 64.7 & 65.3 & 96.0 & 48.8 & 91.3 & 34.4 & 33.2 & 31.6 & 56.2 & 73.4 & 80.2 & 72.0 & 70.5 & 68.9 & 66.1 & 32.4 & 69.6 \\

\texttt{3D\_RoPE} & 61.7 & 2.94 & 65.6 & 65.4 & 95.4 & 46.9 & 91.7 & 34.4 & 33.1 & 31.3 & 55.6 & 74.1 & 79.9 & 67.9 & 69.4 & 70.5 & 64.3 & 33.2 & 70.9 \\

\cellcolor{colorone!25}\texttt{3D\_RoPE\_mixed} & \cellcolor{colorone!25}61.9 & \cellcolor{colorone!25}\textbf{2.53} & \cellcolor{colorone!25}65.1 & \cellcolor{colorone!25}66.3 & \cellcolor{colorone!25}94.6 & \cellcolor{colorone!25}46.3 & \cellcolor{colorone!25}92.1 & \cellcolor{colorone!25}34.4 & \cellcolor{colorone!25}33.4 & \cellcolor{colorone!25}31.7 & \cellcolor{colorone!25}55.7 & \cellcolor{colorone!25}73.4 & \cellcolor{colorone!25}80.0 & \cellcolor{colorone!25}70.5 & \cellcolor{colorone!25}71.5 & \cellcolor{colorone!25}69.4 & \cellcolor{colorone!25}66.3 & \cellcolor{colorone!25}32.3 & \cellcolor{colorone!25}69.6 \\

\end{tabular}}
    \caption{Positional encoding ablations on the v1.2 Base models. Avg. is the simple mean across the displayed tasks. Rank reports the average per-task rank, with lower better.}
    \label{tab:pos-enc-full-no-cropharvest}
\end{table*}

We select \texttt{3D\_RoPE\_mixed}, which uses a joint space-time rotary encoding and obtains the best average rank.

\section{Speedups: Linear vs.\ convolutional patch embedding.} \label{sec:conv_to_linear}
% I think somewhere in this having numbers and cost for doing inference with the new base models on some big region to get per km^2 number difference would be a great way to communicate impact
FlexiViT \cite{beyer2023flexivit} needs a patch embedding that accepts variable patch sizes at runtime;
the natural choice, \texttt{nn.Conv2d} with \texttt{kernel\_size}\,=\,\texttt{stride},
reduces over the input-channel dimension, which in our multi-modal setting is only
$1$--$12$ bands and therefore not 16-byte aligned for \texttt{bf16}. As a result,
cuBLAS dispatches to unvectorized cutlass TensorCore kernels (\texttt{s1688gemm\_\dots\_align1},
$7.5\%$ of GPU time) and prefaces every call with an \texttt{im2col} materialization
that is pure overhead for non-overlapping patches (another $20.2\%$). Replacing the
convolution with a \texttt{reshape + nn.Linear} (mathematically identical, with a
reshape-compatible weight) moves the channel count into the GEMM's reduction
dimension, giving an inner size of $c\!\cdot\!p\!\cdot\!p$ that is naturally aligned
and dispatches to the vectorized cuBLAS TensorCore kernels (\texttt{nvjet\_tst\_*}).
On our 8-GPU FlexiViT-Base setup this drops per-step time from $0.86$\,s to
$0.70$\,s ($1.23\times$ throughput) with no meaningful change to model, loss, or memory.

\section{Improved Finetuning Recipe} \label{app:llrd_ft}

In Section \ref{sec:finetuning_results}, we reuse the finetuning recipe from the OlmoEarth v1 evaluation to compare the pre-trained models in a consistent manner.
However, we find that adopting layer-wise learning rate decay (LLRD) improves finetuning performance for OlmoEarth v1.2 over the frozen start method.
Specifically, we apply LLRD with a layer decay rate of 0.65, where the decoder parameters employ the configured learning rate, the 12th attention block uses 0.65x the learning rate, the 11th block uses $(0.65)^2$, and so on.
We compare the finetuning recipes in Table \ref{tab:finetune_recipe}, referring to the original evaluation recipe as FrozenStart and the improved recipe as LLRD.

\begin{table*}[htpb]
    \centering
    \resizebox{\textwidth}{!}{%
    \begin{tabular}{ll|ccccccccccccccc}

\hdr{} & \hdr{} & \hdr{AWF} & \hdr{AWF} & \hdr{GEA North Africa} & \hdr{Forest Loss Driver} & \hdr{Live Fuel Moisture Content} & \hdr{Live Fuel Moisture Content} & \hdr{Mangrove} & \hdr{Mangrove} & \hdr{Nandi} & \hdr{Nandi} & \hdr{Vessel Detection} & \hdr{Vessel Detection} & \hdr{Vessel Length} & \hdr{Vessel Type} & \hdr{Solar Farm Detection} \\
 & Modalities & S2 & S2, S1 & S2 & S2 & S2 & S2, S1 & S2 & S2, S1 & S2 & S2, S1 & L8 & S1 & S2 & S2 & S2 \\
 & Time series & \ccheck & \ccheck & \ccheck & \ccheck & \ccheck & \ccheck & \ccheck & \ccheck & \ccheck & \ccheck & \ccross & \ccross & \ccross & \ccross & \ccheck \\
Recipe & Metric & Acc. & Acc. & Acc. & Acc. & L1 & L1 & Acc. & Acc. & Acc. & Acc. & F1 & F1 & L1 & Acc. & mIoU \\
\hline
FrozenStart & v1.2 Nano & 75 & 75.5 & 61.5 & \cellcolor{colorone!25} 95.4 & 20.4 & 19.9 & 96.3 & 96.6 & \cellcolor{colorone!25} 79.2 & \cellcolor{colorone!25} 78.1 & \cellcolor{colorone!25} 71.1 & 75.1 & 16.8 & 72.4 & 80.1 \\
FrozenStart & v1.2 Tiny & \cellcolor{colorone!25} 83.5 & \cellcolor{colorone!25} 84.5 & 59.3 & 97.8 & 19.5 & \cellcolor{colorone!25} 18.8 & 97.2 & 97.5 & \cellcolor{colorone!25} 80.7 & \cellcolor{colorone!25} 81.1 & \cellcolor{colorone!25} 75.2 & \cellcolor{colorone!25} 76.9 & \cellcolor{colorone!25} 14.3 & 73.8 & 86.6 \\
FrozenStart & v1.2 Small & 83.0 & 82.0 & 62.0 & \cellcolor{colorone!25} 97.9 & 19.4 & 19.0 & 97.5 & 97.7 & 78.3 & \cellcolor{colorone!25} 80.2 & 75.6 & \cellcolor{colorone!25}  81.9 & 14.2 & 75.9 & 87.5 \\
FrozenStart & v1.2 Base & 83.0 & \cellcolor{colorone!25}  85.0 & 60.2 & 97.1 & 19.1 & 18.3 & 97.5 & 97.6 & \cellcolor{colorone!25}  81.6 & \cellcolor{colorone!25} 82.5 & 74.8 & 78.3 & \cellcolor{colorone!25} 14.5 & \cellcolor{colorone!25} \textbf{75.9} & \cellcolor{colorone!25} 87.3 \\
\hline
LLRD & v1.2 Nano & \cellcolor{colorone!25} 78.0 & \cellcolor{colorone!25} 78.5 & \cellcolor{colorone!25} 66.5 & 93.5 & \cellcolor{colorone!25} 19.6 & \cellcolor{colorone!25} 18.8 & \cellcolor{colorone!25} 97.2 & \cellcolor{colorone!25} 97.5 & 77.5 & 76.8 & 70.7 & 73.8 & \cellcolor{colorone!25} 15.4 & \cellcolor{colorone!25} 73.7 & \cellcolor{colorone!25} 83.1 \\
LLRD & v1.2 Tiny & 82.5 & 82.5 & \cellcolor{colorone!25} 64.7 & \cellcolor{colorone!25} 98.0 & \cellcolor{colorone!25} 19.1 & 18.9 & \cellcolor{colorone!25} 97.8 & \cellcolor{colorone!25} 97.9 & 78.9  & \cellcolor{colorone!25} 81.1 & 74.8 & 75.9 & \cellcolor{colorone!25} 14.3 & \cellcolor{colorone!25} 74.6 & \cellcolor{colorone!25}  88.1 \\
LLRD & v1.2 Small & \cellcolor{colorone!25}  84.5 & \cellcolor{colorone!25} 83.0 & \cellcolor{colorone!25}  63.8 & 97.6 & \cellcolor{colorone!25}  18.2 & \cellcolor{colorone!25}  18.2 & \cellcolor{colorone!25}  97.9 & \cellcolor{colorone!25}  97.9 & \cellcolor{colorone!25}  79.6 & 78.1 & \cellcolor{colorone!25} 76.0 & 76.7 & \cellcolor{colorone!25}  12.8 & \cellcolor{colorone!25}  77.4 & \cellcolor{colorone!25}  89.0 \\
LLRD & v1.2 Base & \cellcolor{colorone!25}  86.0 & 83.5 & \cellcolor{colorone!25}  62.9 & \cellcolor{colorone!25}  97.8 & \cellcolor{colorone!25}  17.7 & \cellcolor{colorone!25}  17.1 & \cellcolor{colorone!25}  98.0 & \cellcolor{colorone!25}  98.0 & 78.9 & 78.2 & \cellcolor{colorone!25}  75.4 & \cellcolor{colorone!25}  80.1 & \cellcolor{colorone!25}  12.3 & \cellcolor{colorone!25}  78.4 & \cellcolor{colorone!25}  90.3 \\

\end{tabular}}
    \caption{Fine-tuning results comparing the FrozenStart method to the layer-wise learning rate decay (LLRD) method. We \hl{highlight} which same-sized models win.}
    \label{tab:finetune_recipe}
\end{table*}

LLRD improves average performance across all three model sizes. For Base, it improves performance on 12 tasks while decreasing performance on 3 tasks. Defining average score as the $\frac{1}{N} (\sum (\textrm{Acc}, \textrm{F1}, \textrm{mIoU}) - \sum(\textrm{L1}))$ (i.e. an average where the L1 tasks -- where lower is better -- are negative), the average scores go from $59.9 \rightarrow 60.9$ for Nano, $62.8 \rightarrow 63.0$ for Tiny, $63.1 \rightarrow 63.5$ for Small and $63.3 \rightarrow 64.0$ for Base. This represents a small but consistent improvement across all model sizes.

\end{document}